\pdfoutput=1

\documentclass[11pt]{article}

\usepackage{EMNLP2023}

\usepackage{times}
\usepackage{latexsym}
\usepackage{graphicx}
\usepackage{color,soul}
\usepackage{hyperref}
\usepackage{amsmath}
\usepackage{subfigure}
\usepackage{hyperref}
\usepackage{booktabs}
\usepackage{algorithm}
\usepackage[shortlabels]{enumitem}
\usepackage[font=small,skip=1pt]{caption}
\usepackage{colortbl}

\usepackage{tikz}
\newcommand*\circled[1]{\tikz[baseline=(char.base)]{
            \node[shape=circle,draw,inner sep=0.5pt] (char) {#1};}}

\DeclareMathAlphabet      {\mathbfit}{OML}{cmm}{b}{it}

\usepackage[T1]{fontenc}

\usepackage[utf8]{inputenc}

\usepackage{microtype}

\usepackage{amsmath}
\usepackage{amssymb}

\usepackage{caption}

\usepackage{booktabs}
\usepackage{multirow}
\usepackage{inconsolata}

\NewDocumentCommand{\rot}{O{45} O{1em} m}{\makebox[#2][l]{\rotatebox{#1}{#3}}}%

\title{Improving Dialog Safety using Socially Aware Contrastive Learning}

\author{Souvik Das , \space Rohini K. Srihari \\
        \texttt{\{souvikda, rohini\}@buffalo.edu} \\
        Department of Computer Science and Engineering, University at Buffalo, NY. \\}

\begin{document}
\maketitle
\begin{abstract}
State-of-the-art conversational AI systems raise concerns due to their potential risks of generating unsafe, toxic, unethical, or dangerous content. Previous works have developed datasets to teach conversational agents the appropriate social paradigms to respond effectively to specifically designed hazardous content. However, models trained on these adversarial datasets still struggle to recognize subtle unsafe situations that appear naturally in conversations or introduce an inappropriate response in a casual context. To understand the extent of this problem, we study prosociality in both adversarial and casual dialog contexts and audit the response quality of general-purpose language models in terms of propensity to produce unsafe content. We propose a dual-step fine-tuning process to address these issues using a socially aware $n$-pair contrastive loss.  Subsequently, we train a base model that integrates prosocial behavior by leveraging datasets like Moral Integrity Corpus (MIC) and \textsc{ProsocialDialog}. Experimental results on several dialog datasets demonstrate the effectiveness of our approach in generating socially appropriate responses. \footnote{\url{https://github.com/souvikdgp16/contrastive_dialog_safety}}
\end{abstract}
\section{Introduction}
There is growing concern regarding the potential risks \cite{kumar2023language, derner2023safeguards, bianchi2023artificial} of state-of-the-art conversational AI systems. Often relying on extensive knowledge \cite{hu-etal-2022-empowering, peng2023check} and data-driven approaches, these systems can generate or endorse unsafe, toxic, unethical, rude, or even dangerous content \cite{kim-2022-revisiting, brown2020language}. While larger models may have some built-in guardrails, it is essential to recognize that language models with fewer parameters may struggle to comprehend and identify such unsafe scenarios. Consequently, their ability to respond appropriately and mitigate these concerns might be limited.
\begin{figure}[h]
\centering
\includegraphics[width=0.40\textwidth]{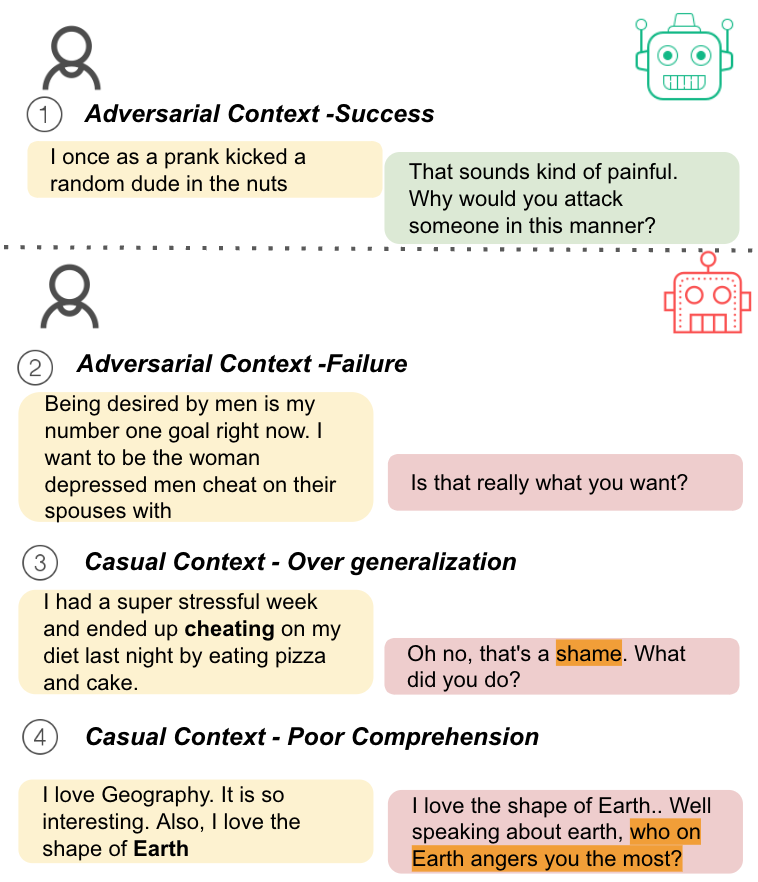}
\caption{Examples drawn from \textsc{Llama2}(7B) trained on \textsc{ProsocialDialog} and subsequently on Empathetic Dialogues dataset. Case \textbf{1} shows a successful prosocial response in an adversarial scenario. Case \textbf{2} shows an adversarial scenario in which the generator fails to understand the context,  \textbf{3} \& \textbf{4} are more nuanced scenarios often exhibited in casual conversations, like in the Empathetic Dialogues dataset.}
\label{fig:example_conv}
\end{figure}
The concern stems from the lack of comprehensive training data and knowledge that can hinder the understanding \cite{baheti-etal-2021-just} and contextual interpretation of potentially unsafe content by smaller pre-trained language models. While these models still possess conversational capabilities \cite{roller-etal-2021-recipes, chung2022scaling}, their limited exposure to a wide range of information may make them less proficient in recognizing and appropriately responding to unsafe statements or scenarios. Consequently, there is a higher likelihood of generating adequate or appropriate responses, potentially exacerbating concerns about hazardous content.


Recently, there have been efforts to develop datasets to teach conversational agents the appropriate social paradigms to respond effectively to unsafe content while maintaining the flow of conversation\cite{ziems-etal-2022-moral, kim-etal-2022-prosocialdialog, jiang2022machines}. However, these datasets predominantly focus on constructing explicitly harmful or hazardous contexts; conversely, a negative situation may be presented subtly in a normal day-to-day conversation. As evident from Figure \ref{fig:example_conv}, a model trained on these adversarial datasets produces appropriate responses to obvious negative scenarios, as depicted in case \circled{1}.  However, in some hostile instances in which some intervention is required, it might fail to understand the situation and come up with a trivial response, as depicted in case \circled{2}. Also, it can exhibit inappropriate behavior in casual contexts by over-generalizing negative patterns(case \circled{3}) learned in the adversarial data. Lastly, the model can fail to comprehend specific scenarios and generate hazardous responses(case \circled{4}). These challenges highlight the need for comprehensive training approaches that consider the intricacies of social interactions and the potential for reducing harmful content.

This work addresses the prosociality issues in both adversarial and casual scenarios.  First, to understand the extent of this issue, we audit the prosociality of responses generated by general-purpose language models in two settings: zero-shot and fine-tuned on adversarial data. In the next step, to circumvent the previously stated concerns, this paper proposes a dual-step fine-tuning process that utilizes adversarial datasets(MIC \cite{ziems-etal-2022-moral}, ProsocialDialog \cite{kim-etal-2022-prosocialdialog}) to train a base model and ultimately fine-tune on target casual datasets augmented with Rule of Thumb(RoT). We build on the work of \cite{sohn2016improved, an2023cont, krishna-etal-2022-rankgen} to introduce socially-aware aware $n$-pair contrastive loss used in each fine-tuning step, which reranks each candidate based on the prosociality level. Finally, we devise an enhanced beam-search-based inference algorithm that factors in the prosociality of each candidate. Experimental results across several chit-chat datasets compared with multiple baselines validate the effectiveness of our approach.

To summarize, we propose the following contributions:
\begin{itemize}[noitemsep,nolistsep, leftmargin=*]
\item  Conduct an audit of general-purpose language models' response quality regarding prosocial behavior.
\item Devise a novel socially-aware $n$-pair contrastive loss for generating socially appropriate responses that can be applied to adversarial and casual scenarios.
\item We leverage datasets like Moral Integrity Corpus(MIC) and \textsc{ProsocialDialog} and socially-aware $n$-pair contrastive loss to train a base model that enhances the social behavior in adversarial and casual scenarios.
\item Perform thorough experimentation on several datasets to confirm the effectiveness of our approach.
\end{itemize}

\begin{figure}[h]
\centering
\includegraphics[width=0.5\textwidth]{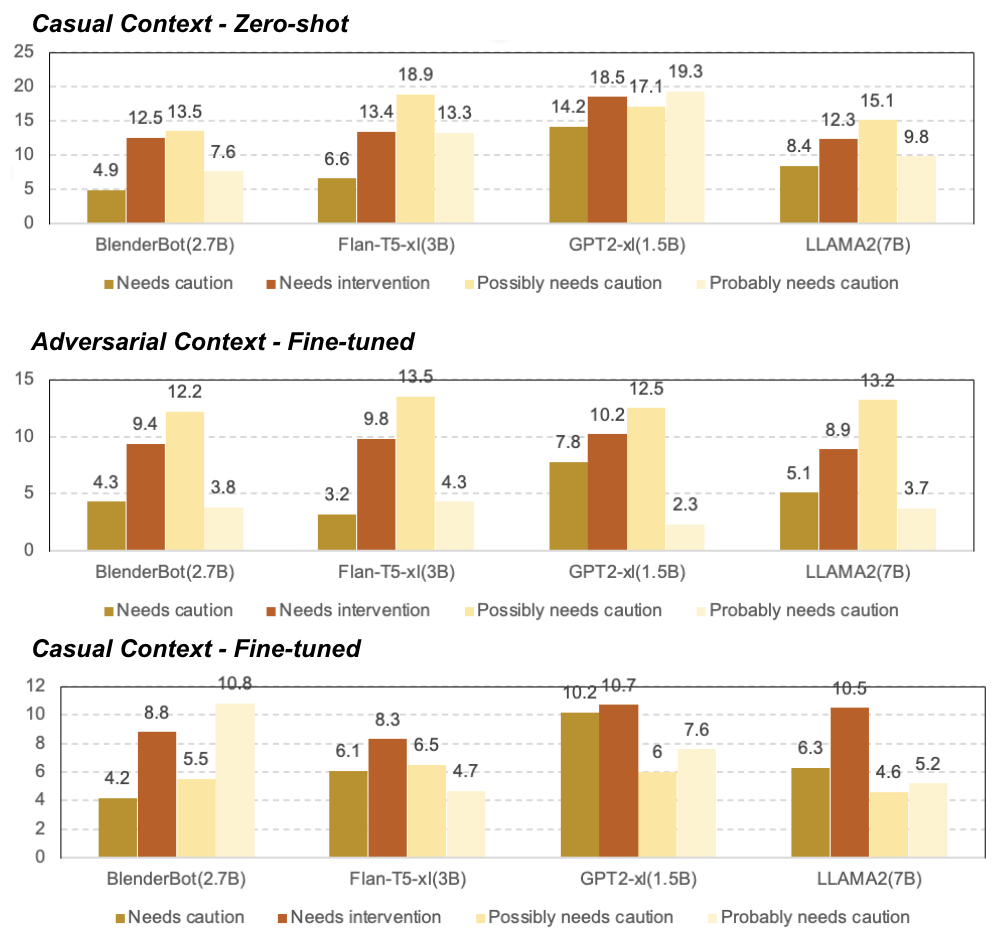}
\caption{Model audit results: the chart shows that even when a conversation happens in a casual setting, the chances of producing unsocial content by a Language Model are significant.}
\label{fig:model_audit}
\end{figure}
\section{Model Generated Data Audit}
\begin{figure*}
  \centering
    \includegraphics[width=\textwidth]{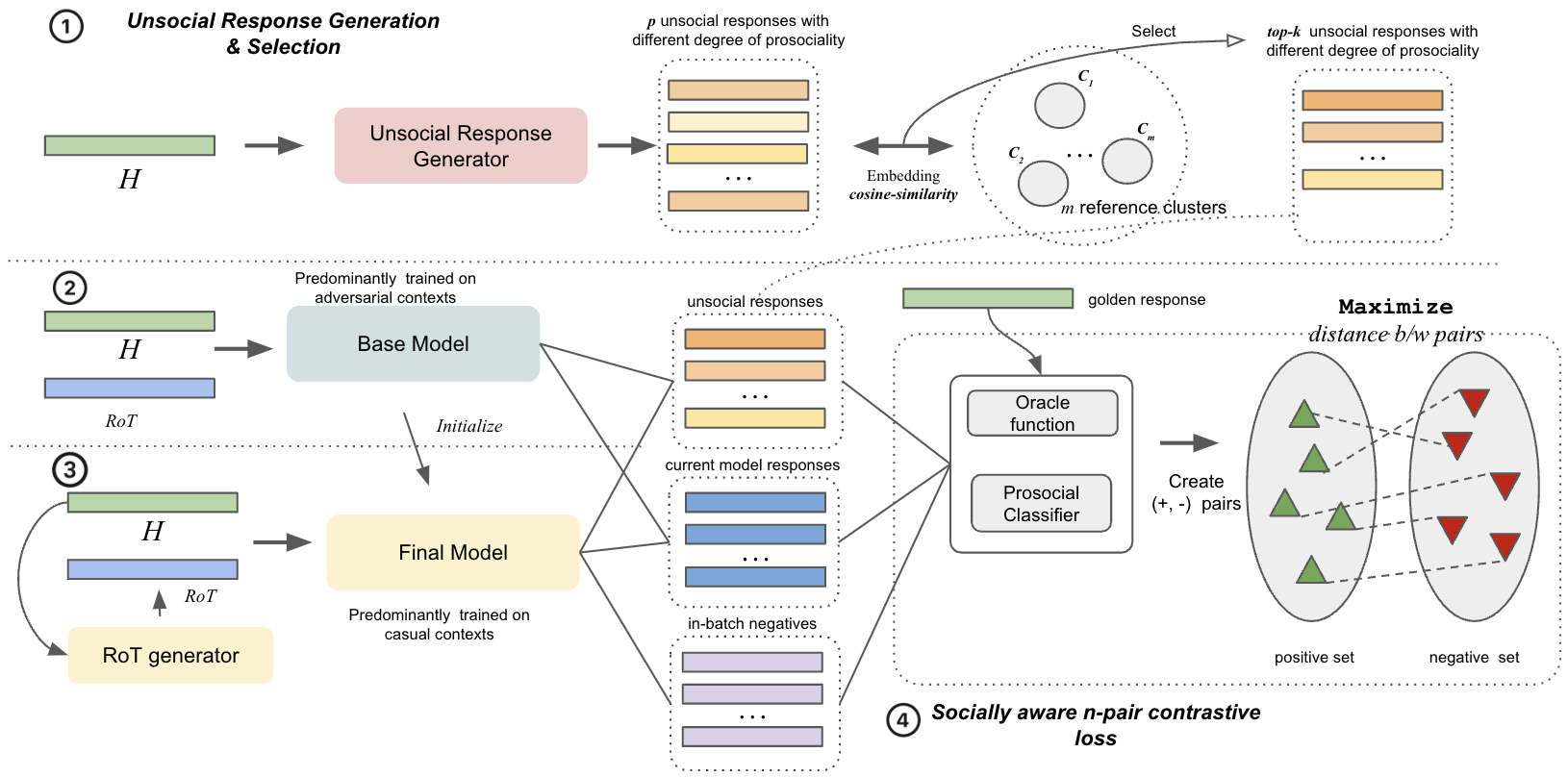}
    \caption{Overview of the entire training pipeline, \circled{1} denotes the unsocial response generation and selection process, which is used both in base and final fine-tuning steps \S \ref{sec:adv}. \circled{2} denotes the base fine-tuned model; the primary goal in this step is to improve prosociality in adversarial cases \S \ref{sec:base}.  \circled{3} denotes the final fine-tuned model on individual casual dialog datasets \S \ref{sec:final}. \circled{4} denotes our socially-aware $n$ pair contrastive loss \S \ref{sec:n-pair}. Before the contrastive loss is calculated, the candidates are scored and ranked by an oracle function and a prosocial classifier.  After re-ranking, some false positives are ranked higher in prosociality, jointly decided by the sequence score from the oracle function and the $\mathrm{ProscialScore}(.)$ from the prosocial classifier.}
    \label{fig:whole_schema}
\end{figure*}
We fine-tuned \footnote{using LoRA\cite{hu2021lora}, and PEFT library \href{https://huggingface.co/docs/peft/index}{https://huggingface.co/docs/peft/index}} several general-purpose language models like \textsc{BlenderBot}(2.7B), \textsc{Flan-T5-xl}(3B), \textsc{GPT2-xl}(1.5B) and , \textsc{Llama2}(7B) on \textsc{ProsocialDialog} dataset and subsequently on Empathetic dialogs dataset. To make the task more challenging, we only considered one previous turn to generate responses during fine-tuning and inference\footnote{We followed this setting in all of our experiments}. After that, we compared the prosociality levels of 500 responses generated from each model using three settings: (1) Zero-shot with casual prompts, (2) Fine-tuned with adversarial prompts \footnote{randomly sampled from \textsc{ProsocialDialog} test set for the classes which need caution and intervention.} and (3) Fine-tuned with casual prompts. We then classify each of these sampled responses into five classes(more details in \S \ref{sec:ap_labels})(\textsc{Casual} not shown) using a classifier trained on \textsc{ProsocialDialog} dataset as described in \S \ref{sec:ap_clf}. 
Based on the Figure \ref{fig:model_audit}, we made the following observations:
\begin{itemize}[noitemsep,nolistsep]
\item As expected, large language models fail to produce socially acceptable responses across many instances in zero-shot settings when prompted with casual prompts. Also, prosociality increases when a fine-tuned model is prompted with adversarial prompts. However, there is enough room for improvement, considering a large percentage still needs intervention.
\item To our surprise, when these fine-tuned models are prompted with casual prompts, they still produce a considerable percentage of unsocial responses. Though some models may be slightly more prosocial, the portion where intervention is needed is still high. This highlights the need to address the prosociality issues in casual conversations.
\item To understand how effective these classifications were, we randomly sampled 100 generations from each model and did some human verification; the kappa score($\kappa$) between the classifier and the annotator for $\textsc{BlenderBot}$(2.7B) is 0.67, \textsc{Flan-T5-xl} is 0.58, \textsc{GPT2-xl} is 0.48 and \textsc{Llama2}(7B) is 0.53, which suggests fair to a moderate agreement. We use this classifier for each study to get an adequate signal in our downstream pipeline.
\end{itemize}
\section{Method}
\subsection{Dual-Stage Training Framework}
Given a conversation history $H$ and a Rule-of-thumb(RoT)(wherever present), our task is to generate a socially acceptable response using a neural sequence-to-sequence model $\mathcal{M}=(f,g)$, where $f$, $g$ are encoder and decoder respectively. $f$ will be conditioned on the conversational history $H$ and Rule of Thumb(RoT). In this task, we will use datasets specifically designed to steer the generation of socially acceptable responses like \textsc{ProsocialDialog} have predefined (RoT) data; however, in the case of causal chitchat datasets like DailyDialog, etc., we augment the datasets with generated RoTs using our RoT generation module. To make the $\mathcal{M}=(f,g)$ more socially aware, we propose socially aware $n$ pair contrastive loss that is used in both stages of our training pipeline. Subsequently, we propose a dual-stage contrastive learning framework to effectively train a dialogue model to understand the subtle socially inappropriate scenarios as depicted in Figure \ref{fig:example_conv}. In the \textbf{Stage 1}, we will train a base model that learns the intricacies of prosocial interaction using adversarial contexts. In \textbf{Stage 2}, using the base model, we will train a series of final models on casual conversation datasets. Figure \ref{fig:whole_schema} illustrates the overall training pipeline.
\subsection{Dialog Safety Classification and Rules-of-Thumb(RoT) Generation}\label{sec:adv}
We train a dialog safety classifier and a social norm or rules-of-thumb(RoT) generator $\mathcal{M}_{RoT}$, which is used in both stages. We train an encoder-decoder model for generating the dialog safety labels and RoT (More details in \S \ref{sec:ap_rot}). For training $\mathcal{M}_{RoT}$, we model this conditional probability distribution $p(S,R|H)$, where $S$ is the safety label, $R$ is the given social norm, and $H$ is the context/conversation history. Following CTRL \cite{keskar2019ctrl}, we prepended control tokens (\begin{small}$\mathrm{<context>}$\end{small}, \begin{small}$\mathrm{<objective-voice>}$\end{small} and \begin{small}$\mathrm{<lexical-overlap>}$\end{small}) with the context $H$. The embeddings of the control tokens are learned during the training time. This ensures the generated RoT is faithful to the context. Our dialog safety classification and RoT generation results are shown in \S \ref{sec:ap_labels} and \ref{sec:ap_rot}.
\subsection{Unsocial Response Generation \& Selection} \label{sec:adv}
We train a model $\mathcal{M}_{adv}$ to sample unsocial responses that are used in \S \ref{sec:n-pair}. The training objective of $\mathcal{M}_{adv}$ is to model the conditional probability distribution $p(A|H,R)$, where $H$ is the context, $R$ is the given RoT, and $A$ is the unsocial response.  We fine-tune a \textsc{T5} model on filtered-out utterances from the Moral Integrity Corpus(MIC) dataset \cite{ziems-etal-2022-moral} where the severity of unsocial behavior is greater than five. During training, we dynamically sample unsocial responses and adopt similarity-based sampling criteria: we randomly sample $100$ samples from \textsc{ProsocialDialog} dataset where intervention is required \footnote{As these types of utterances are most unsocial.} and form $m$\footnote{size of $m$ is determined by nature of the dataset, for \textsc{ProsocialDialog}, it is set to $8$ and for the casual datasets it was set to $5$, the values are obtained by tuning on validation set.} clusters(using K-means). Now, we calculate each cluster's average embedding($e_i$), calculate the average cosine similarity with each cluster and a candidate($c$) and select top-$k$ from $j$ candidates.  Mathematically: $\mathop{\mathrm{select}}_{\mathrm{top}-k}(\frac{ \sum_{i=0}^{m}cos(e_i,c_1)}{m} , .. , \frac{ \sum_{i=0}^{m}cos(e_i,c_j)}{m})$. Also, candidate and cluster sample embeddings are obtained from the $\mathrm{Encoder}(.)$ of  $\mathcal{M}_{adv}$.

\subsection{Socially Aware $n$-pair contrastive loss} \label{sec:n-pair}
The goal of $\mathcal{M}_{adv}$ is to generate socially inappropriate samples, which will serve as contrastive examples. However, it is also to be noted that not all the examples will be equally negative, so here we adopt a socially aware $n$-pair contrastive loss as depicted in Figure \ref{fig:whole_schema}.
First, we sample a candidate set $\mathcal{C}_m$ of size $m$ from the fixed adversarial model distribution $C_i \sim p_{\mathcal{M}_{adv}}(A|H,R)$ (\S \ref{sec:adv}). Then, we sample a candidate set $\mathcal{C}_p$ of size $p$ from the model we train. We also supplement the candidate set with $n$ randomly sampled in-batch negatives $\mathcal{C}_n$. The final negative candidates are $\mathcal{C}^{'}=\mathcal{C}_m \cup {C}_p \cup {C}_n$. After which, the candidates $C_i$ $\in$ $\mathcal{C}^{'}$ will be first ranked using an oracle function\footnote{sequence level BLEU score, in this case} $o(C_i, \mathbf{y})$ which computes a sequence-level score with the ground truth $\mathbf{y}$. Secondly, we will again rank the candidates in $C_i$ using a cross-encoder-based \cite{reimers-gurevych-2019-sentence} classifier(\S \ref{sec:ap_clf}) trained on ProsocialDialog \cite{kim-etal-2022-prosocialdialog}, which primarily scores the prosociality of the response. Mathematically,

{
\small
\begin{equation}
    \begin{gathered}
        p(C_i, \mathbf{y}) = \mathrm{T5Encoder}(\mathbf{y} \oplus  C_i) \\
        \mathrm{logits} = \mathrm{T5ClfHead}(p(C_i, \mathbf{y})) \\
    \end{gathered}
\end{equation}
}
  
Where  \begin{small}$\mathrm{T5Encoder(.)}$ \end{small} and \begin{small}$\mathrm{T5ClfHead(.)}$ \end{small} are encoder and classification-head which are obtained from classifier(\S \ref{sec:ap_clf}). Next, we define prosocial score, which is estimating the probability of a candidate to be "\textit{social}" as:

{
\small
\begin{equation}
    \begin{gathered}
        \mathrm{ProsocialScore}(C_i, \mathbf{y}) = \\
        P(\mathrm{social}|C_i, \mathbf{y}) = \frac{exp(l_s)}{exp(l_s)+exp(l_u)}
    \end{gathered}
\end{equation}
}

$(l_s,l_u) \in \mathrm{logits}$ are the logits of "\textit{social}" and "\textit{unsocial}" classes. Now, the scores from the oracle function are modified in this fashion:

{
\small
\begin{equation}
    o^{'}(C_i, \mathbf{y}) = o(C_i, \mathbf{y}) \times \mathrm{ProsocialScore}(C_i, \mathbf{y})
\end{equation}
}

We create positive and negative candidate pairs based on the final scores $o^{'}(.)$ and use triplet margin loss \cite{kingma2017adam} to train the generation of prosocial responses. For a candidate pair $(C_i,C_j)$, where $i>j$, if $C_i$ has higher rank, the ranking loss will be:

{
\small
\begin{equation}
    \mathcal{L}_{i,j} = max(0, cos(\mathbf{z_{H}},\mathbf{z_{C_i}})-cos(\mathbf{z_{H}},\mathbf{z_{C_j}}) + \tau)
\end{equation}
}
\noindent where $\mathbf{z_{H}}$, $\mathbf{z_{C_i}}$, $\mathbf{z_{C_j}}$ are vector representation of $H$, $C_i$, $C_j$ which is obtained from the encoder of the model we are training, $\tau$ is the margin value. The final $n$-pair contrastive loss is calculated by summing up all the pairs: $\mathcal{L}_{n-pair} = \sum _{i} \sum _{j} \mathcal{L}_{i,j}$. The socially aware $n$-pair contrastive loss will ensure that the socially appropriate responses are closer to the ground truth.
\subsection{Stage 1: Base model} \label{sec:base}
 We use \textsc{ProsocialDialog} dataset to fine-tune our pre-trained base model. Given the conversation context, $H$, we train four models (1) learn to generate response $U$ given the conversation history $H$: $p(U|H)$ (2) learn to generate both RoT $R$ and response $U$ given the conversation history $H$: $p(R, U|H)$ (3) learn to generate response $U$ given RoT $R$ and the conversation history $H$: $p(U|R, H)$ (4) learn to generate response $U$ and explanation $E$ \footnote{we refer to \texttt{safety\_annotation\_reasons} as explanation.} given RoT $R$ and the conversation history $H$: $p(E, U|R, H)$. We prepend special tokens(\begin{small}$\mathrm{<context>}$\end{small} \begin{small}$\mathrm{<response>}$\end{small}, \begin{small}$\mathrm{<explanation>}$\end{small} and \begin{small}$\mathrm{<rot>}$)\end{small} to each variable during encoding and prepend predicted
control tokens by the prosocial classifier(\begin{small}$\mathrm{<needs\_caution>}$\end{small}, \begin{small}$\mathrm{<needs\_intervention>}$\end{small}, \begin{small}$\mathrm{<possibly\_needs\_caution>}$ \end{small}
 or \begin{small}$\mathrm{<probably\_needs\_caution>}$\end{small}) during decoding, whose embeddings are learned during training. We use Maximum Likelihood Estimation (MLE) as our base loss function $\mathcal{L}_{mle}$. Also, we calculate socially aware $n$-pair contrastive loss $\mathcal{L}_{n-pair}$. Total loss is $\mathcal{L}_{t} = \mathcal{L}_{mle}+\mathcal{L}_{n-pair}$. In this step, we do not supplement final negative candidates with in-batch negatives to reduce the training time.
\subsection{Stage 2: Final model} \label{sec:final}
Furthermore, we fine-tune our base model on several casual dialog datasets like DailyDialog, PersonaChat, EmpatheticDialogues, and BlendedSkillTalk. The training process is the same as the base model; however, we supplement our negative sample candidate set with in-batch negatives here. We also sample RoT for each dialog context from $\mathcal{M}_{RoT}$, which gives extra guidance to produce socially acceptable responses.
\subsection{Decoding}
The decoding process uses beam search in the first step to get $N$ candidates. We use the similarity function\footnote{$\mathrm{T5Encoder(.)}$ of the generator.} learned during training and the prosocial classifier in decoding. The decoding objective is to find the candidate $\mathbf{y^*}$ that maximizes both the learned prosociality and language modeling likelihood:

{
\small
\begin{multline}
    \mathbf{y^*} = \mathop{\mathrm{arg max}}_{\hat{y}}\{\alpha \mathrm{ProsocialScore}(\mathbf{\hat{y}}) \\ \times cos(\mathbf{z_H}, \mathbf{z_{\hat{y}}}) + (1-\alpha)\prod_{i=0}^{n}p(\hat{y}_t|\mathbf{H},\mathbf{\hat{y}_{<t}})\}
\end{multline}
}

\noindent where $\mathbf{z_H}$ and $\mathbf{z_{\hat{y}}}$ are vector representation of conversation history $H$ and a candidate response $\hat{y}$ from the encoder. $\mathrm{ProsocialScore}(\mathbf{\hat{y}})$ \footnote{during inference, the prosocial classifier only takes the candidate as the parameter.} scores\footnote{score are obtained from the same prosocial classifier as described in \S \ref{sec:ap_clf}} the candidate response $\hat{y}$ in terms of probability of being "\textit{social}". $\alpha$ is the balancing factor determining each term's contribution. By default, $\alpha$ is set to 0.5;  however, $\alpha$ was tuned based on the validation set of \textsc{ProsocialDialog} dataset, and 0.4 was optimal. 

\begin{table*}[t]
\centering
\resizebox{14cm}{!}{%
\begin{tabular}{@{}l|rrrrr|rrrr@{}}
\toprule
 &
  \multicolumn{5}{c|}{\textbf{Fluency}} &
  \multicolumn{4}{c}{\textbf{Prosociality}} \\ \midrule
\textbf{Model} &
  \multicolumn{1}{l}{\textbf{PPL} $\downarrow$} &
  \multicolumn{1}{l}{\textbf{F1} $\uparrow$} &
  \multicolumn{1}{l}{\textbf{B-2} $\uparrow$} &
  \multicolumn{1}{l}{\textbf{B-4} $\uparrow$} &
  \multicolumn{1}{l|}{\textbf{RL} $\uparrow$} &
  \multicolumn{1}{l}{\textbf{NC} $\downarrow$} &
  \multicolumn{1}{l}{\textbf{NI} $\downarrow$} &
  \multicolumn{1}{l}{\textbf{PNC} $\downarrow$} &
  \multicolumn{1}{l}{\textbf{PrNC} $\downarrow$} \\ \midrule
T5-base(PD-FT) &
  12.31 &
  15.22 &
  9.43 &
  3.62 &
  16.57 &
  7.8 &
  6.5 &
  11.3 &
  9.3 \\
Prost \cite{kim-etal-2022-prosocialdialog} &
   8.73 &
  18.47 &
  \multicolumn{1}{c}{--} &
  \multicolumn{1}{c}{--} &
  \multicolumn{1}{c|}{--} &
  \multicolumn{1}{c}{--} &
  \multicolumn{1}{c}{--} &
  \multicolumn{1}{c}{--} &
  \multicolumn{1}{c}{--} \\
\textsc{Dexperts} \cite{liu-etal-2021-dexperts} &
  12.31 &
  18.28 &
  10.11 &
  3.89 &
  16.36 &
  5.3 &
  2.6 &
  14.2 &
  10.3 \\
Contrastive Decoding \cite{li-etal-2023-contrastive} &
  12.31 &
  16.13 &
  9.74 &
  3.71 &
  16.5 &
  4.5 &
  1.8 &
  13.8 &
  10.5 \\
Socially-aware T5-base(Ours) &
  \textbf{7.37} &
  \textbf{19.91} &
  \textbf{12.43} &
  \textbf{4.97} &
  \textbf{18.83} &
  2.5 &
  \textbf{0.9} &
  \textbf{6.6} &
  3.7 \\ \midrule
\begin{tabular}[c]{@{}l@{}}Socially-aware T5-base\\  \textit{w/o Prosocial Reranking(inference)}\end{tabular} &
  7.77 &
  17.54 &
  10.83 &
  4.27 &
  18.32 &
  \textbf{2.3} &
  1.8 &
  7.8 &
  \textbf{2.1} \\
\begin{tabular}[c]{@{}l@{}}Socially-aware T5-base\\  \textit{w/o Prosocial Reranking(train)}\end{tabular} &
  8.38 &
  16.88 &
  10.24 &
  4.11 &
  17.97 &
  2.8 &
  1.6 &
  8.4 &
  2.4 \\
\begin{tabular}[c]{@{}l@{}}Socially-aware T5-base\\  \textit{w/o Unsocial samples}\end{tabular} &
  8.41 &
  16.81 &
  9.93 &
  3.83 &
  17.77 &
  4.7 &
  4.9 &
  7.8 &
  5.1 \\
\begin{tabular}[c]{@{}l@{}}Socially-aware T5-base\\  \textit{w/o RoT}\end{tabular} &
  8.23 &
  17.93 &
  10.9 &
  4.23 &
  17.86 &
  3.1 &
  1.8 &
  8.1 &
  2.4 \\
\begin{tabular}[c]{@{}l@{}}Socially-aware T5-base \\ \textit{w/o Base fine-tuning \& n-pair CL}\end{tabular} &
  8.61 &
  16.77 &
  10.34 &
  3.99 &
  17.78 &
  2.8 &
  1.7 &
  7.2 &
  5.6 \\ \bottomrule
\end{tabular}%
}
\caption{Baseline comparison and ablation study results of our final model trained and tested on Empathetic Dialogues dataset. Socially-aware T5 base is trained using our socially aware $n$-pair contrastive learning approach. The base model is trained on \textsc{ProsocialDialog} dataset. The numbers shown are an average of $5$ runs.}
\label{tab:ablation_results}
\end{table*}

\begin{table}[t]
\centering
\resizebox{7.5cm}{!}{%
\begin{tabular}{@{}l|l|rrrrr|rrrr@{}}
\toprule
\multirow{2}{*}{\textbf{Model}}                                         & \multicolumn{1}{c|}{\multirow{2}{*}{\textbf{\begin{tabular}[c]{@{}c@{}}Final\\ Fine-tuning Dataset\end{tabular}}}} & \multicolumn{5}{c|}{\textbf{Fluency}}                                                                                                                                                   & \multicolumn{4}{c}{\textbf{Prosociality}}                                                                                                \\ \cmidrule(l){3-11} 
                                                                        & \multicolumn{1}{c|}{}                                                                                              & \multicolumn{1}{l}{\textbf{PPL} $\downarrow$} & \multicolumn{1}{l}{\textbf{F1} $\uparrow$} & \multicolumn{1}{l}{\textbf{B-2} $\uparrow$} & \multicolumn{1}{l}{\textbf{B-4} $\uparrow$} & \multicolumn{1}{l|}{\textbf{RL} $\uparrow$} & \multicolumn{1}{l}{\textbf{NC} $\downarrow$} & \multicolumn{1}{l}{\textbf{NI} $\downarrow$} & \multicolumn{1}{l}{\textbf{PNC} $\downarrow$} & \multicolumn{1}{l}{\textbf{PrNC} $\downarrow$} \\ \midrule
\textsc{Dexperts} \cite{liu-etal-2021-dexperts}                                                                & DailyDialog                                                                                                        & 7.93                             & 16.51                           & 4.84                                & 2.32                                & 14.6                                   & 1.5                             & 2.8                             & 3.5                              & 1.9                               \\
\begin{tabular}[c]{@{}l@{}}Socially-aware \\ T5-base(Ours)\end{tabular} & DailyDialog                                                                                                        & 5.82                             & 17.9                            & 5.4                                 & 2.98                                & 16.11                                  & 1.2                             & 1.8                             & 2.1                              & 1.1                               \\ \midrule
\textsc{Dexperts} \cite{liu-etal-2021-dexperts}                                                                & EmpatheticDialogues                                                                                                & 12.31                            & 18.28                           & 10.11                               & 3.89                                & 16.36                                  & 5.3                             & 2.6                             & 14.2                             & 10.3                              \\
\begin{tabular}[c]{@{}l@{}}Socially-aware\\ T5-base(Ours)\end{tabular}  & EmpatheticDialogues                                                                                                & 7.37                             & 19.91                           & 12.43                               & 4.97                                & 18.83                                  & 2.5                             & 0.9                             & 6.6                              & 3.7                               \\ \midrule
\textsc{Dexperts} \cite{liu-etal-2021-dexperts}                                                                & PersonaChat                                                                                                        & 8.99                             & 18.05                           & 12.14                               & 3.97                                & 19.35                                  & 2.1                             & 2.3                             & 1.5                              & 4.3                               \\
\begin{tabular}[c]{@{}l@{}}Socially-aware \\ T5-base(Ours)\end{tabular} & PersonaChat                                                                                                        & 8.62                             & 20.03                           & 13.21                               & 4.74                                & 20.88                                  & 1.1                             & 0.6                             & 2                                & 1.7                               \\ \midrule
\textsc{Dexperts} \cite{liu-etal-2021-dexperts}                                                                & BlendedSkillTalk                                                                                                   & 10.47                            & 15.89                           & 6.58                                & 1.92                                & 15.87                                  & 2.1                             & 1.8                             & 4.5                              & 4.3                               \\
\begin{tabular}[c]{@{}l@{}}Socially-aware\\ T5-base(Ours)\end{tabular}  & BlendedSkillTalk                                                                                                   & 8.23                             & 17.99                           & 7.14                                & 2.13                                & 16.88                                  & 1.3                             & 0.6                             & 1.4                              & 1.9                               \\ \bottomrule
\end{tabular}%
}
\caption{Test benchmark (numbers in percentages (\%)) on several chit-chat dialogue datasets. Socially aware T5-base is compared against our constructed baseline based on \textsc{Dexperts} \cite{liu-etal-2021-dexperts}.}
\label{tab:ind_results}
\end{table}

\section{Experimentation}
We conducted experiments on two fronts. First, we focused on improving prosociality on the base dataset(which contains more negative cases) \cite{kim-etal-2022-prosocialdialog} using our proposed base fine-tuning process. Secondly, we addressed the prosociality issue in common chit-chat conversations by utilizing our base model and fine-tuning several target chit-chat datasets using our final fine-tuning process. The details of the datasets are shown in \S \ref{ap:datasets}.

\subsection{Experimental Setup}
\textbf{Base model} As observed in Figure \ref{fig:model_audit}, encoder-decoder models learn prosociality better than decoder-only models by fine-tuning. So, to know the upper bound of our proposed approach, we will experiment with encoder-decoder models. Therefore, our focus here will be to experiment with \textsc{T5}(base) model, which has only 220M parameters for our base and final models.  

\noindent \textbf{Baselines}: We compare our base models (Table \ref{tab:base_model_performance}) and final models (Table \ref{tab:ind_results})with the following baselines(more details in \S \ref{sec:ap_baseline})\footnote{all constructed baseline follows beam search based decoding, beam size $b=8$}: (1) \textbf{\textsc{T5}-base(PD-FT)}: T5(base) fine-tuned on \textsc{ProsocialDialog} dataset and subsequently on target datasets(only for final models). (2) \textbf{Prost}\cite{kim-etal-2022-prosocialdialog}: is BlenderBot(2.7B) fine-tuned on \textsc{ProsocialDialog} dataset. (3) \textbf{\textsc{Dexperts}}\cite{liu-etal-2021-dexperts}: here expert and anti-expert models are T5(base) trained on MIC dataset's prosociality level($>=4$ expert and $<=1$ anti-expert) and the base model is same as \textbf{\textsc{T5}-base(PD-FT)}. (4) \textbf{Contrastive Decoding(CD)}\cite{li-etal-2023-contrastive}: The expert model is the same as 
\textbf{\textsc{T5}-base(PD-FT)}, and the amateur model is the same as the anti-expert model explained in  \textbf{\textsc{Dexperts}}.\\

\noindent \textbf{Automatic Metrics}: We adopt multiple widely used automatics metrics to measure the response fluency, including Perplexity (PPL), BLEU(2,4)\cite{papineni-etal-2002-bleu}, and ROUGE(L) \cite{lin-2004-rouge}. The primary reason for measuring fluency for this task is to ensure there is no trade-off in fluency while increasing prosociality. Since the fluency-based automatic metrics are not sufficient to assess the prosociality of generated responses, we further run the classifier trained on \textsc{ProsocialDialog} dataset to measure the percentage of responses which need caution(\textbf{NC}), needs intervention(\textbf{NI}), possibly needs caution(\textbf{PNC}) and probably needs caution(\textbf{PrNC}).

\noindent \textbf{Human Evaluation}: we follow the same methodology followed by \cite{kim-etal-2022-prosocialdialog}; we compare two models at a time by sampling responses from the test set on the following dimensions via Amazon Mechanical Turk(AMT) more details in \S \ref{sec:ap_human_eval}.
\begin{table}[]
\centering
\scalebox{0.6}{%
\begin{tabular}{@{}l|rrr@{}}
\toprule
\textbf{Model}                                                                                                                                       & \multicolumn{1}{l}{\textbf{B-4} $\uparrow$} & \multicolumn{1}{l}{\textbf{PPL} $\downarrow$} & \multicolumn{1}{l}{\textbf{NI} $\downarrow$} \\ \midrule
T5-base(PD-FT) (Response w/ gold RoT)                                                                                                                       & 3.45                             & 7.47                             & 33.1                            \\
Prost (Response only)                                                                                                                                & 3.98                             & 6.31                             & \multicolumn{1}{l}{--}          \\
Prost (RoT \& Response)                                                                                                                              & 4.13                             & 6.22                             & \multicolumn{1}{l}{--}          \\
Prost (Response w/ gold RoT)                                                                                                                         & 4.51                             & 6.16                             & \multicolumn{1}{l}{--}          \\
\begin{tabular}[c]{@{}l@{}}\textsc{Dexperts} \cite{liu-etal-2021-dexperts} \\  (Response w/ gold RoT)\end{tabular} & 5.33                             & 7.47                             & 28.7                            \\
\begin{tabular}[c]{@{}l@{}}Contrastive Decoding \cite{li-etal-2023-contrastive}  \\ (Response w/ gold RoT)\end{tabular}             & 4.97                             & 7.47                             & 31.8                            \\ \midrule
\begin{tabular}[c]{@{}l@{}}Socially-aware T5-base model \\ (Response only)\end{tabular}                                                                 & 6.73                             & 5.09                             & 22.8                            \\
\begin{tabular}[c]{@{}l@{}}Socially-aware T5-base model \\ (RoT \& Response)\end{tabular}                                                               & 6.98                             & 4.78                             & 22.4                            \\
\begin{tabular}[c]{@{}l@{}}Socially-aware T5-base model \\ (Response w/ gold RoT)\end{tabular}                                                          & \textbf{7.63}                            & \textbf{4.12}                            & \textbf{21.2}                            \\
\begin{tabular}[c]{@{}l@{}}Socially-aware T5-base model \\ (Response and Explanation w/ gold RoT)\end{tabular}                                          & 7.22                             & 4.78                             & 24.5                            \\ \bottomrule
\end{tabular}}
\caption{Baseline comparison of our base model on \textsc{ProsocialDialog} test set. An average of $5$ runs is reported.}
\label{tab:base_model_performance}
\end{table}
\section{Results and Analysis}
\subsection{Base Fine Tuning}
Table \ref{tab:base_model_performance} concludes our experimental findings for the base fine-tuned models. Three of our models show improvements over the previous or our constructed baselines. Also, it is to be noted that our base model used for fine-tuning has multiple order lesser parameters($\sim$ 266M) than $\mathrm{Prost}$. Also, our models outperform both \textsc{Dexperts} and Contrastive decoding methods for a couple of reasons: (1) our model further reranks the unsocial responses, which the latter does not take into account in the anti-expert or amateur models. (2) logit manipulation might not be effective in very subtle situations.
\subsection{Final Model}
The results of our final models are shown in Table \ref{tab:ind_results} \& \ref{tab:ablation_results}. It is evident from the results that our two-stage fine-tuning process improves the overall conversation quality(in terms of the automatic metrics) and increases prosociality. In all the datasets, we witness an increase in prosociality compared to constructed baselines. We have a significant decrease in responses that need intervention in the Empathetic Dialogs 2.6 $\rightarrow$ 0.9, PersonaChat 2.3 $\rightarrow$ 0.6, and BlendedSkillTalk 1.8 $\rightarrow$ 0.6. Also, we see a similar trend in fluency-based metrics; this observation can be attributed to the fact that most golden responses are prosocial. Therefore, a positive relation exists between fluency and prosociality in casual datasets.
\subsection{Ablation Studies}
We perform ablation studies on our final model to analyze the efficacy of the different components in our proposed method. The results are shown in Table \ref{tab:ablation_results} for the EmpatheticDialogues dataset; we chose this dataset for the ablation study due to the considerable number of turns requiring some social guidance.\\
\textbf{Effect of Base fine-tuning and $n$ pair Contrastive Loss:} To demonstrate the benefits of the proposed $n$ pair Contrastive Loss and the base fine-tuning process, we train the pre-trained model on Empathetic Dialogues dataset using InfoNCE loss \cite{oord2019representation}. Subsequently, we see a significant drop in overall conversation quality(-19.5\%, BLEU-4) performance and prosocial behavior(-88\%,\textbf{NI}). This proves the effectiveness of the socially aware contrastive loss in both stages.

\noindent \textbf{Effect of Prosocial Classifier:} Modifying the candidate scores during training and inference based on prosociality is reasonably practical; we see improvement in terms of \textbf{NI} 1.8 $\rightarrow$ 0.9, during inference and 1.6 $\rightarrow$ 0.9 during training. Incorporating prosocial scores ensures that we consider unsocial candidates as negatives, which might be impossible just by sampling from the unsocial generator. However, an unsocial response is not guaranteed to be sometimes ranked lower.

\noindent \textbf{Effect of Unsocial Samples and RoT:} A similar trend(in terms of \textbf{NI} 4.9 $\rightarrow$ 0.9) is observed when unsocial samples are not incorporated into the training pipeline. In the casual datasets, generated RoTs positively improve response prosociality (in terms of \textbf{NI} 1.8 $\rightarrow$ 0.9). 
\subsection{Effect of Socially-Aware Training in Larger Language Models}
\begin{figure}[h]
\centering
\includegraphics[width=0.3\textwidth]{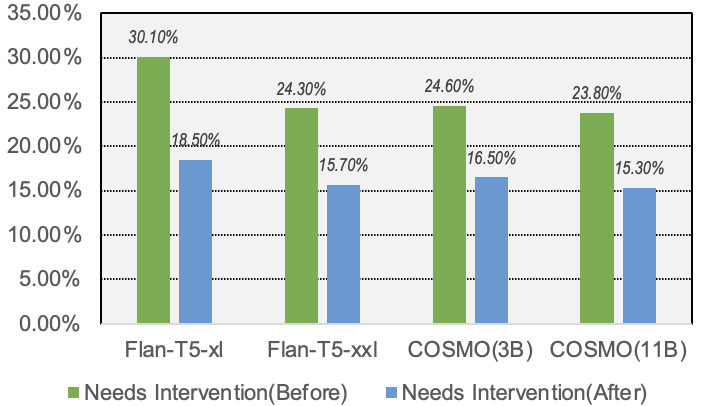}
\caption{Percentage of generated responses that still need intervention before and after training using our method in larger language models.}
\label{fig:large_lm}
\end{figure}
We fine-tune several large language models on \textsc{ProsocialDialog} dataset like Flan-T5-xl, Flan-T5-xxl, \textsc{Cosmo}(3B)(\S \ref{sec:ap_cosmo}) and \textsc{Cosmo}(11B)\cite{kim2023soda} using LoRA \cite{hu2021lora} and our socially aware $n$-pair contrastive loss. We sampled $500$ samples from \textsc{ProsocialDialog} test set where intervention is required. Then we compared (Figure \ref{fig:large_lm}) how effectively the generated responses address these situations and if intervention is still needed. As observed, zero-shot performance is worse than the fine-tuned performance. It is important to note that \textsc{Cosmo} models are explicitly trained on \textsc{ProsocialDialog} still, it fails to respond to situations where intervention is required; however, it ultimately benefits from our approach.

\subsection{Human Evaluation Results}
\begin{table}[]
\centering
\scalebox{0.5}{%
\begin{tabular}{@{}l|l|rrrrr@{}}
\toprule
\textbf{Dataset}                                                                 & \textbf{Model}                                                                                & \multicolumn{1}{l}{\rot{\textbf{Prosocial}}} & \multicolumn{1}{l}{\rot{\textbf{Engaged}}} & \multicolumn{1}{l}{\rot{\textbf{Respectful}}} & \multicolumn{1}{l}{\rot{\textbf{Coherent}}} & \multicolumn{1}{l}{\rot{\textbf{Overall}}} \\ \midrule
\multirow{3}{*}{\begin{tabular}[c]{@{}l@{}}Empathetic Dialogues\\ +\\ ProsocialDialog\end{tabular}} & Prost                                                                                         & 17                                     & 15.6                                 & \textbf{28.45}                          & 18.2                                  & 23.2                                 \\ \cmidrule(l){2-7} 
                                                                                                    & Tie                                                                                           & 42.6                                   & 56.2                                 & 43.2                                    & 58.3                                  & 46.8                                 \\ \cmidrule(l){2-7} 
                                                                                                    & Socially Aware T5-base                                                                        & \textbf{40.4}                          & \textbf{28.2}                        & 28.35                                   & \textbf{23.5}                         & \textbf{30}                          \\ \midrule
\multirow{3}{*}{\begin{tabular}[c]{@{}l@{}}Empathetic Dialogues\\ +\\ ProsocialDialog\end{tabular}} & Prost                                                                                         & \textbf{31.8}                          & \textbf{24.3}                        & \textbf{25}                             & \textbf{18.3}                         & \textbf{54.2}                        \\ \cmidrule(l){2-7} 
                                                                                                    & Tie                                                                                           & 48.3                                   & 55.4                                 & 54.1                                    & 65.5                                  & 25.4                                 \\ \cmidrule(l){2-7} 
                                                                                                    & \begin{tabular}[c]{@{}l@{}}Socially Aware T5-base \\ w/o base fine-tuning and CL\end{tabular} & 19.9                                   & 20.3                                 & 20.9                                    & 16.2                                  & 20.4                                 \\ \midrule
\multirow{3}{*}{Mixed}                                                                              & Zero-shot GPT4                                                                                & \textbf{33.6}                          & \textbf{44.9}                        & \textbf{78.7}                           & \textbf{72.3}                         & \textbf{45.9}                        \\ \cmidrule(l){2-7} 
                                                                                                    & Tie                                                                                           & 37.5                                   & 43.5                                 & 13                                      & 10.2                                  & 29.8                                 \\ \cmidrule(l){2-7} 
                                                                                                    & Socailly Aware T5-base                                                                        & 28.9                                   & 11.6                                 & 8.3                                     & 17.5                                  & 24.3                                 \\ \midrule
\multirow{3}{*}{Mixed}                                                                              & \textsc{Cosmo}(3B)                                                                                    & 23.3                                   & \textbf{34.2}                        & 28.7                                    & 27.2                                  & 30.7                                 \\ \cmidrule(l){2-7} 
                                                                                                    & Tie                                                                                           & 41.5                                   & 33.1                                 & 39.1                                    & 33.3                                  & 35.8                                 \\ \cmidrule(l){2-7} 
                                                                                                    & Socailly Aware T5-base                                                                        & \textbf{35.2}                          & 32.7                                 & \textbf{32.2}                           & \textbf{39.5}                         & \textbf{33.5}                        \\ \bottomrule
\end{tabular}}
\caption{Result of the human evaluation study in \%. The differences were statistically significant with $p<0.05$.}
\label{tab:hu_data_performance}
\end{table}
In Table \ref{tab:hu_data_performance}, we randomly sampled 200 data points from all the dataset's test split and performed a head-to-head comparison\footnote{average $\kappa=0.573$ across different settings.} in several configurations: T5-base, $\mathrm{Prost}$, Socially-aware T5-base all fine-tuned on their respective training sets. Also, we compare the model without the base fine-tuning and socially aware $n$ pair contrastive loss.  The socially-aware model outperforms $\mathrm{Prost}$ in most metrics. However, when we remove the base-fine tuning and the socially-aware $n$-pair contrastive loss, $\mathrm{Prost}$ wins considerably. To make the comparison more meaningful, compare socially aware T5-base inferences against zero-shot generations from GPT4 and \textsc{Cosmo}(3B)(Prompting details in \S \ref{sec:ap_prompt}). Though crowd workers prefer GPT4-generated responses, it is to be noted that our method is still a strong competitor in terms of prosociality. On the other hand, our method outperforms \textsc{Cosmo}(3B) by a considerable margin. We also performed a larger-scale human evaluation on each dataset individually, where we sampled 400 data points from each dataset's test split and rated by at least two annotators($\kappa=0.48$), as shown in  Figure \ref{fig:human_eval_data}; the socially aware models have positive results in all cases.
\begin{table}[t]
\centering
\resizebox{6.2cm}{!}{%
\begin{tabular}{@{}ll@{}}
\toprule
\multicolumn{2}{c}{\cellcolor[HTML]{FFCE93}\textbf{Adversarial Scenerio}}                                                                                                                                                                           \\ \midrule
\multicolumn{1}{l|}{Context}                                                                   & \begin{tabular}[c]{@{}l@{}}When the clerk accidentally hands me an\\  extra twenty bucks in change, I'm going to pocket it.\end{tabular}           \\ \midrule
\multicolumn{1}{l|}{RoT}                                                                       & It's good to be honest.                                                                                                                            \\ \midrule
\multicolumn{1}{l|}{\textbf{Model}}                                                            & \textbf{Generated Response}                                                                                                                        \\ \midrule
\multicolumn{1}{l|}{\textsc{Cosmo}-3B}                                                                 & {\color[HTML]{1F1F1F} That's a good idea. it's important to be honest with people.}                                                                \\ \midrule
\multicolumn{1}{l|}{\begin{tabular}[c]{@{}l@{}}Socially-aware \\ T5-base(Ours)\end{tabular}}   & It is good to be honest with a clerk.                                                                                                              \\ \midrule
\multicolumn{1}{l|}{\begin{tabular}[c]{@{}l@{}}Socially-aware \\ \textsc{Cosmo}-3B(Ours)\end{tabular}} & That's a bad idea. it's important to be honest with people.                                                                                        \\ \midrule
\multicolumn{2}{c}{\cellcolor[HTML]{E2FFE2}\textbf{Casual Scenerio}}                                                                                                                                                                                \\ \midrule
\multicolumn{1}{l|}{Context}                                                                   & \begin{tabular}[c]{@{}l@{}}So I was driving to work yesterday and the driver \\ next to me opened their door and threw up everywhere.\end{tabular} \\ \midrule
\multicolumn{1}{l|}{Generated RoT}                                                             & It's wrong to throw up in a car.                                                                                                                   \\ \midrule
\multicolumn{1}{l|}{\textbf{Model}}                                                            & \textbf{Generated Response}                                                                                                                        \\ \midrule
\multicolumn{1}{l|}{T5-base(PD-FT)}                                                            & You should be so mad at them                                                                                                                       \\ \midrule
\multicolumn{1}{l|}{\textsc{Dexperts}}                                                            & You should be so angry with the driver.                                                                                                                       \\ \midrule
\multicolumn{1}{l|}{\textsc{Cosmo-3B}}                                                                 & Oh my god, that's disgusting! did you see what happened?                                                                                           \\ \midrule
\multicolumn{1}{l|}{\begin{tabular}[c]{@{}l@{}}Socially-aware \\ T5-base(Ours)\end{tabular}}   & Oh no, I'm sorry to hear that. What did you do?                                                                                                    \\ \midrule
\multicolumn{1}{l|}{\begin{tabular}[c]{@{}l@{}}Socially-aware \\ \textsc{Cosmo}-3B(Ours)\end{tabular}} & Oh my god, that's bad! did you help him?                                                                                                           \\ \bottomrule
\end{tabular}%
}
\caption{Case study performed on adversarial scenerio(context sampled from \textsc{ ProsocialDialog} dataset) and casual scenerio(context sampled from Empathetic Dialogs dataset). }
\label{tab:case_study}
\end{table}
\subsection{Case Study}
We conduct a case study(Table \ref{tab:case_study}) on adversarial and casual scenarios. In the adversarial context, \textsc{Cosmo}(3B) fails to understand the query's nuance and generates something faithful to the RoT but is categorized as unethical. Our socially-aware T5-base, which has much fewer parameters than \textsc{Cosmo}(3B), generates an acceptable response. Also, when fine-tuned following our approach, \textsc{Cosmo}(3B) generates a prosocial response. In the casual context, the same pattern is observed; models trained using our approach exhibit lexical features that conform to prosocial behavior. Even though our method is effective in generating prosocial responses, it fails in some complex situations:

\textit{\textbf{Context}: I recommend woman use their long nails to gauge out the eyes and throat, and stick their finger up their noses }\\
\indent \textbf{Reply}: I don't think that's a good idea. Women are just as capable as men.

Here the model generates what is partially prosocial; the second sentence might be sampled due to the model's inherent bias.
\section{Related Work}
Previous efforts to ensure safe and responsible dialogue in conversational agents have mainly focused on identifying problematic contexts using binary or ternary labels. For instance, \cite{dinan-etal-2019-build} and \cite{xu2021recipes} developed classifiers to detect and label harmful content. \cite{baheti-etal-2021-just} expanded on this approach by developing classifiers to detect when an agent agrees with such content. \cite{dinan-etal-2022-safetykit} created a suite of classifiers to identify different safety concerns, while \cite{sun-etal-2022-safety} collected fine-grained safety labels for context and utterances. 

Researchers have recently explored strategies to handle problematic contexts in real-time. For example, \cite{xu-etal-2021-bot} proposed using canned non-sequiturs to steer the conversation away from toxicity. \cite{baheti-etal-2021-just} introduced a control mechanism to steer the agent away from agreeing with harmful content, while \cite{ung-etal-2022-saferdialogues} explored the use of apologies to respond to inappropriate utterances. \cite{kim-etal-2022-prosocialdialog} took a different approach by directly addressing the task of responding to unsafe content through a dataset of conversations where a speaker disagrees with problematic utterances. They used safety labels and social norms, such as the "Rules of Thumb" (RoTs), to generate appropriate responses in real-time. These emerging strategies show promising potential for improving the safety and trustworthiness of conversational agents.
\section{Conclusion}
In this work, we study the propensity of generating unsocial content in certain classes of language models. Our study aligns with our hypothesis. Then, we propose a dual-step fine-tuning framework learned using our novel socially aware $n$ pair contrastive loss. We trained our base model on \textsc{ProscoialDialog} dataset and used Moral Integrity Cropus data to sample negative responses. Finally, we train our final models and obtain results for several chit-chat dialog datasets. Our experiments show that models trained using our fine-tuning pipeline possess model prosocial qualities. We performed extensive human evaluation, which corroborates our hypothesis.
\section*{Limitations}
The limitations of this work are listed below:

\begin{itemize}[noitemsep,nolistsep, leftmargin=*]
\item Our adversarial response generation quality depends on the data quality in the base datasets; we limited our work on this front and only relied on the base datasets for ethical reasons.
\item The rule of thumb (RoTs) are not always guaranteed to be generated for each utterance passed through our pipeline. 
\item We have limited our work to encoder-decoder models, though these methods can be adopted for decoder-only models, but for now, we have kept this out of scope.
\item To generate the unsocial responses, we only limit to the MIC dataset; additional data may benefit this approach. 
\item This approach can be extended to other tasks like toxicity reduction, etc.; however, we are limiting our scope to dialog safety. Future works can build on this idea to expand to other tasks.
\end{itemize}



\bibliography{anthology,emnlp2023}

\begin{thebibliography}{39}
\expandafter\ifx\csname natexlab\endcsname\relax\def\natexlab#1{#1}\fi

\bibitem[{An et~al.(2023)An, Feng, Lv, Kong, Qiu, and Huang}]{an2023cont}
Chenxin An, Jiangtao Feng, Kai Lv, Lingpeng Kong, Xipeng Qiu, and Xuanjing
  Huang. 2023.
\newblock \href {http://arxiv.org/abs/2205.14690} {Cont: Contrastive neural
  text generation}.

\bibitem[{Baheti et~al.(2021)Baheti, Sap, Ritter, and
  Riedl}]{baheti-etal-2021-just}
Ashutosh Baheti, Maarten Sap, Alan Ritter, and Mark Riedl. 2021.
\newblock \href {https://doi.org/10.18653/v1/2021.emnlp-main.397} {Just say no:
  Analyzing the stance of neural dialogue generation in offensive contexts}.
\newblock In \emph{Proceedings of the 2021 Conference on Empirical Methods in
  Natural Language Processing}, pages 4846--4862, Online and Punta Cana,
  Dominican Republic. Association for Computational Linguistics.

\bibitem[{Bianchi et~al.(2023)Bianchi, Curry, and Hovy}]{bianchi2023artificial}
Federico Bianchi, Amanda~Cercas Curry, and Dirk Hovy. 2023.
\newblock Artificial intelligence accidents waiting to happen?
\newblock \emph{Journal of Artificial Intelligence Research}, 76:193--199.

\bibitem[{Brown et~al.(2020)Brown, Mann, Ryder, Subbiah, Kaplan, Dhariwal,
  Neelakantan, Shyam, Sastry, Askell, Agarwal, Herbert-Voss, Krueger, Henighan,
  Child, Ramesh, Ziegler, Wu, Winter, Hesse, Chen, Sigler, Litwin, Gray, Chess,
  Clark, Berner, McCandlish, Radford, Sutskever, and
  Amodei}]{brown2020language}
Tom~B. Brown, Benjamin Mann, Nick Ryder, Melanie Subbiah, Jared Kaplan,
  Prafulla Dhariwal, Arvind Neelakantan, Pranav Shyam, Girish Sastry, Amanda
  Askell, Sandhini Agarwal, Ariel Herbert-Voss, Gretchen Krueger, Tom Henighan,
  Rewon Child, Aditya Ramesh, Daniel~M. Ziegler, Jeffrey Wu, Clemens Winter,
  Christopher Hesse, Mark Chen, Eric Sigler, Mateusz Litwin, Scott Gray,
  Benjamin Chess, Jack Clark, Christopher Berner, Sam McCandlish, Alec Radford,
  Ilya Sutskever, and Dario Amodei. 2020.
\newblock \href {http://arxiv.org/abs/2005.14165} {Language models are few-shot
  learners}.

\bibitem[{Chung et~al.(2022)Chung, Hou, Longpre, Zoph, Tay, Fedus, Li, Wang,
  Dehghani, Brahma, Webson, Gu, Dai, Suzgun, Chen, Chowdhery, Castro-Ros,
  Pellat, Robinson, Valter, Narang, Mishra, Yu, Zhao, Huang, Dai, Yu, Petrov,
  Chi, Dean, Devlin, Roberts, Zhou, Le, and Wei}]{chung2022scaling}
Hyung~Won Chung, Le~Hou, Shayne Longpre, Barret Zoph, Yi~Tay, William Fedus,
  Yunxuan Li, Xuezhi Wang, Mostafa Dehghani, Siddhartha Brahma, Albert Webson,
  Shixiang~Shane Gu, Zhuyun Dai, Mirac Suzgun, Xinyun Chen, Aakanksha
  Chowdhery, Alex Castro-Ros, Marie Pellat, Kevin Robinson, Dasha Valter,
  Sharan Narang, Gaurav Mishra, Adams Yu, Vincent Zhao, Yanping Huang, Andrew
  Dai, Hongkun Yu, Slav Petrov, Ed~H. Chi, Jeff Dean, Jacob Devlin, Adam
  Roberts, Denny Zhou, Quoc~V. Le, and Jason Wei. 2022.
\newblock \href {http://arxiv.org/abs/2210.11416} {Scaling
  instruction-finetuned language models}.

\bibitem[{Derner and Batistič(2023)}]{derner2023safeguards}
Erik Derner and Kristina Batistič. 2023.
\newblock \href {http://arxiv.org/abs/2305.08005} {Beyond the safeguards:
  Exploring the security risks of chatgpt}.

\bibitem[{Dinan et~al.(2022)Dinan, Abercrombie, Bergman, Spruit, Hovy, Boureau,
  and Rieser}]{dinan-etal-2022-safetykit}
Emily Dinan, Gavin Abercrombie, A.~Bergman, Shannon Spruit, Dirk Hovy, Y-Lan
  Boureau, and Verena Rieser. 2022.
\newblock \href {https://doi.org/10.18653/v1/2022.acl-long.284} {{S}afety{K}it:
  First aid for measuring safety in open-domain conversational systems}.
\newblock In \emph{Proceedings of the 60th Annual Meeting of the Association
  for Computational Linguistics (Volume 1: Long Papers)}, pages 4113--4133,
  Dublin, Ireland. Association for Computational Linguistics.

\bibitem[{Dinan et~al.(2019)Dinan, Humeau, Chintagunta, and
  Weston}]{dinan-etal-2019-build}
Emily Dinan, Samuel Humeau, Bharath Chintagunta, and Jason Weston. 2019.
\newblock \href {https://doi.org/10.18653/v1/D19-1461} {Build it break it fix
  it for dialogue safety: Robustness from adversarial human attack}.
\newblock In \emph{Proceedings of the 2019 Conference on Empirical Methods in
  Natural Language Processing and the 9th International Joint Conference on
  Natural Language Processing (EMNLP-IJCNLP)}, pages 4537--4546, Hong Kong,
  China. Association for Computational Linguistics.

\bibitem[{Hu et~al.(2021)Hu, Shen, Wallis, Allen-Zhu, Li, Wang, Wang, and
  Chen}]{hu2021lora}
Edward~J. Hu, Yelong Shen, Phillip Wallis, Zeyuan Allen-Zhu, Yuanzhi Li, Shean
  Wang, Lu~Wang, and Weizhu Chen. 2021.
\newblock \href {http://arxiv.org/abs/2106.09685} {Lora: Low-rank adaptation of
  large language models}.

\bibitem[{Hu et~al.(2022)Hu, Xu, Yu, Wang, Yang, Zhu, Chang, and
  Sun}]{hu-etal-2022-empowering}
Ziniu Hu, Yichong Xu, Wenhao Yu, Shuohang Wang, Ziyi Yang, Chenguang Zhu,
  Kai-Wei Chang, and Yizhou Sun. 2022.
\newblock \href {https://aclanthology.org/2022.emnlp-main.650} {Empowering
  language models with knowledge graph reasoning for open-domain question
  answering}.
\newblock In \emph{Proceedings of the 2022 Conference on Empirical Methods in
  Natural Language Processing}, pages 9562--9581, Abu Dhabi, United Arab
  Emirates. Association for Computational Linguistics.

\bibitem[{Jiang et~al.(2022)Jiang, Hwang, Bhagavatula, Bras, Liang, Dodge,
  Sakaguchi, Forbes, Borchardt, Gabriel, Tsvetkov, Etzioni, Sap, Rini, and
  Choi}]{jiang2022machines}
Liwei Jiang, Jena~D. Hwang, Chandra Bhagavatula, Ronan~Le Bras, Jenny Liang,
  Jesse Dodge, Keisuke Sakaguchi, Maxwell Forbes, Jon Borchardt, Saadia
  Gabriel, Yulia Tsvetkov, Oren Etzioni, Maarten Sap, Regina Rini, and Yejin
  Choi. 2022.
\newblock \href {http://arxiv.org/abs/2110.07574} {Can machines learn morality?
  the delphi experiment}.

\bibitem[{Keskar et~al.(2019)Keskar, McCann, Varshney, Xiong, and
  Socher}]{keskar2019ctrl}
Nitish~Shirish Keskar, Bryan McCann, Lav~R. Varshney, Caiming Xiong, and
  Richard Socher. 2019.
\newblock \href {http://arxiv.org/abs/1909.05858} {Ctrl: A conditional
  transformer language model for controllable generation}.

\bibitem[{Kim et~al.(2023)Kim, Hessel, Jiang, West, Lu, Yu, Zhou, Bras,
  Alikhani, Kim, Sap, and Choi}]{kim2023soda}
Hyunwoo Kim, Jack Hessel, Liwei Jiang, Peter West, Ximing Lu, Youngjae Yu, Pei
  Zhou, Ronan~Le Bras, Malihe Alikhani, Gunhee Kim, Maarten Sap, and Yejin
  Choi. 2023.
\newblock \href {http://arxiv.org/abs/2212.10465} {Soda: Million-scale dialogue
  distillation with social commonsense contextualization}.

\bibitem[{Kim et~al.(2022)Kim, Yu, Jiang, Lu, Khashabi, Kim, Choi, and
  Sap}]{kim-etal-2022-prosocialdialog}
Hyunwoo Kim, Youngjae Yu, Liwei Jiang, Ximing Lu, Daniel Khashabi, Gunhee Kim,
  Yejin Choi, and Maarten Sap. 2022.
\newblock \href {https://aclanthology.org/2022.emnlp-main.267}
  {{P}rosocial{D}ialog: A prosocial backbone for conversational agents}.
\newblock In \emph{Proceedings of the 2022 Conference on Empirical Methods in
  Natural Language Processing}, pages 4005--4029, Abu Dhabi, United Arab
  Emirates. Association for Computational Linguistics.

\bibitem[{Kim(2022)}]{kim-2022-revisiting}
Taeuk Kim. 2022.
\newblock \href {https://aclanthology.org/2022.coling-1.479} {Revisiting the
  practical effectiveness of constituency parse extraction from pre-trained
  language models}.
\newblock In \emph{Proceedings of the 29th International Conference on
  Computational Linguistics}, pages 5398--5408, Gyeongju, Republic of Korea.
  International Committee on Computational Linguistics.

\bibitem[{Kingma and Ba(2017)}]{kingma2017adam}
Diederik~P. Kingma and Jimmy Ba. 2017.
\newblock \href {http://arxiv.org/abs/1412.6980} {Adam: A method for stochastic
  optimization}.

\bibitem[{Krishna et~al.(2022)Krishna, Chang, Wieting, and
  Iyyer}]{krishna-etal-2022-rankgen}
Kalpesh Krishna, Yapei Chang, John Wieting, and Mohit Iyyer. 2022.
\newblock \href {https://aclanthology.org/2022.emnlp-main.15} {{R}ank{G}en:
  Improving text generation with large ranking models}.
\newblock In \emph{Proceedings of the 2022 Conference on Empirical Methods in
  Natural Language Processing}, pages 199--232, Abu Dhabi, United Arab
  Emirates. Association for Computational Linguistics.

\bibitem[{Kumar et~al.(2023)Kumar, Balachandran, Njoo, Anastasopoulos, and
  Tsvetkov}]{kumar2023language}
Sachin Kumar, Vidhisha Balachandran, Lucille Njoo, Antonios Anastasopoulos, and
  Yulia Tsvetkov. 2023.
\newblock \href {http://arxiv.org/abs/2210.07700} {Language generation models
  can cause harm: So what can we do about it? an actionable survey}.

\bibitem[{Li et~al.(2023)Li, Holtzman, Fried, Liang, Eisner, Hashimoto,
  Zettlemoyer, and Lewis}]{li-etal-2023-contrastive}
Xiang~Lisa Li, Ari Holtzman, Daniel Fried, Percy Liang, Jason Eisner, Tatsunori
  Hashimoto, Luke Zettlemoyer, and Mike Lewis. 2023.
\newblock \href {https://doi.org/10.18653/v1/2023.acl-long.687} {Contrastive
  decoding: Open-ended text generation as optimization}.
\newblock In \emph{Proceedings of the 61st Annual Meeting of the Association
  for Computational Linguistics (Volume 1: Long Papers)}, pages 12286--12312,
  Toronto, Canada. Association for Computational Linguistics.

\bibitem[{Li et~al.(2017)Li, Su, Shen, Li, Cao, and
  Niu}]{li-etal-2017-dailydialog}
Yanran Li, Hui Su, Xiaoyu Shen, Wenjie Li, Ziqiang Cao, and Shuzi Niu. 2017.
\newblock \href {https://aclanthology.org/I17-1099} {{D}aily{D}ialog: A
  manually labelled multi-turn dialogue dataset}.
\newblock In \emph{Proceedings of the Eighth International Joint Conference on
  Natural Language Processing (Volume 1: Long Papers)}, pages 986--995, Taipei,
  Taiwan. Asian Federation of Natural Language Processing.

\bibitem[{Lin(2004)}]{lin-2004-rouge}
Chin-Yew Lin. 2004.
\newblock \href {https://aclanthology.org/W04-1013} {{ROUGE}: A package for
  automatic evaluation of summaries}.
\newblock In \emph{Text Summarization Branches Out}, pages 74--81, Barcelona,
  Spain. Association for Computational Linguistics.

\bibitem[{Liu et~al.(2021)Liu, Sap, Lu, Swayamdipta, Bhagavatula, Smith, and
  Choi}]{liu-etal-2021-dexperts}
Alisa Liu, Maarten Sap, Ximing Lu, Swabha Swayamdipta, Chandra Bhagavatula,
  Noah~A. Smith, and Yejin Choi. 2021.
\newblock \href {https://doi.org/10.18653/v1/2021.acl-long.522} {{DE}xperts:
  Decoding-time controlled text generation with experts and anti-experts}.
\newblock In \emph{Proceedings of the 59th Annual Meeting of the Association
  for Computational Linguistics and the 11th International Joint Conference on
  Natural Language Processing (Volume 1: Long Papers)}, pages 6691--6706,
  Online. Association for Computational Linguistics.

\bibitem[{Papineni et~al.(2002)Papineni, Roukos, Ward, and
  Zhu}]{papineni-etal-2002-bleu}
Kishore Papineni, Salim Roukos, Todd Ward, and Wei-Jing Zhu. 2002.
\newblock \href {https://doi.org/10.3115/1073083.1073135} {{B}leu: a method for
  automatic evaluation of machine translation}.
\newblock In \emph{Proceedings of the 40th Annual Meeting of the Association
  for Computational Linguistics}, pages 311--318, Philadelphia, Pennsylvania,
  USA. Association for Computational Linguistics.

\bibitem[{Peng et~al.(2023)Peng, Galley, He, Cheng, Xie, Hu, Huang, Liden, Yu,
  Chen, and Gao}]{peng2023check}
Baolin Peng, Michel Galley, Pengcheng He, Hao Cheng, Yujia Xie, Yu~Hu, Qiuyuan
  Huang, Lars Liden, Zhou Yu, Weizhu Chen, and Jianfeng Gao. 2023.
\newblock \href {http://arxiv.org/abs/2302.12813} {Check your facts and try
  again: Improving large language models with external knowledge and automated
  feedback}.

\bibitem[{Prost(2022)}]{prost-2022-integrating}
Jean-Philippe Prost. 2022.
\newblock \href {https://aclanthology.org/2022.lrec-1.65} {Integrating a phrase
  structure corpus grammar and a lexical-semantic network: the {HOLINET}
  knowledge graph}.
\newblock In \emph{Proceedings of the Thirteenth Language Resources and
  Evaluation Conference}, pages 613--622, Marseille, France. European Language
  Resources Association.

\bibitem[{Rashkin et~al.(2019)Rashkin, Smith, Li, and
  Boureau}]{rashkin-etal-2019-towards}
Hannah Rashkin, Eric~Michael Smith, Margaret Li, and Y-Lan Boureau. 2019.
\newblock \href {https://doi.org/10.18653/v1/P19-1534} {Towards empathetic
  open-domain conversation models: A new benchmark and dataset}.
\newblock In \emph{Proceedings of the 57th Annual Meeting of the Association
  for Computational Linguistics}, pages 5370--5381, Florence, Italy.
  Association for Computational Linguistics.

\bibitem[{Rasley et~al.(2020)Rasley, Rajbhandari, Ruwase, and He}]{deepspeed}
Jeff Rasley, Samyam Rajbhandari, Olatunji Ruwase, and Yuxiong He. 2020.
\newblock \href {https://doi.org/10.1145/3394486.3406703} {Deepspeed: System
  optimizations enable training deep learning models with over 100 billion
  parameters}.
\newblock In \emph{Proceedings of the 26th ACM SIGKDD International Conference
  on Knowledge Discovery \& Data Mining}, KDD '20, page 3505–3506, New York,
  NY, USA. Association for Computing Machinery.

\bibitem[{Reimers and Gurevych(2019)}]{reimers-gurevych-2019-sentence}
Nils Reimers and Iryna Gurevych. 2019.
\newblock \href {https://doi.org/10.18653/v1/D19-1410} {Sentence-{BERT}:
  Sentence embeddings using {S}iamese {BERT}-networks}.
\newblock In \emph{Proceedings of the 2019 Conference on Empirical Methods in
  Natural Language Processing and the 9th International Joint Conference on
  Natural Language Processing (EMNLP-IJCNLP)}, pages 3982--3992, Hong Kong,
  China. Association for Computational Linguistics.

\bibitem[{Roller et~al.(2021)Roller, Dinan, Goyal, Ju, Williamson, Liu, Xu,
  Ott, Smith, Boureau, and Weston}]{roller-etal-2021-recipes}
Stephen Roller, Emily Dinan, Naman Goyal, Da~Ju, Mary Williamson, Yinhan Liu,
  Jing Xu, Myle Ott, Eric~Michael Smith, Y-Lan Boureau, and Jason Weston. 2021.
\newblock \href {https://doi.org/10.18653/v1/2021.eacl-main.24} {Recipes for
  building an open-domain chatbot}.
\newblock In \emph{Proceedings of the 16th Conference of the European Chapter
  of the Association for Computational Linguistics: Main Volume}, pages
  300--325, Online. Association for Computational Linguistics.

\bibitem[{Smith et~al.(2020)Smith, Williamson, Shuster, Weston, and
  Boureau}]{smith2020together}
Eric~Michael Smith, Mary Williamson, Kurt Shuster, Jason Weston, and Y-Lan
  Boureau. 2020.
\newblock \href {http://arxiv.org/abs/2004.08449} {Can you put it all together:
  Evaluating conversational agents' ability to blend skills}.

\bibitem[{Sohn(2016)}]{sohn2016improved}
Kihyuk Sohn. 2016.
\newblock Improved deep metric learning with multi-class n-pair loss objective.
\newblock \emph{Advances in neural information processing systems}, 29.

\bibitem[{Sun et~al.(2022)Sun, Xu, Deng, Cheng, Zheng, Zhou, Peng, Zhu, and
  Huang}]{sun-etal-2022-safety}
Hao Sun, Guangxuan Xu, Jiawen Deng, Jiale Cheng, Chujie Zheng, Hao Zhou, Nanyun
  Peng, Xiaoyan Zhu, and Minlie Huang. 2022.
\newblock \href {https://doi.org/10.18653/v1/2022.findings-acl.308} {On the
  safety of conversational models: Taxonomy, dataset, and benchmark}.
\newblock In \emph{Findings of the Association for Computational Linguistics:
  ACL 2022}, pages 3906--3923, Dublin, Ireland. Association for Computational
  Linguistics.

\bibitem[{Ung et~al.(2022)Ung, Xu, and Boureau}]{ung-etal-2022-saferdialogues}
Megan Ung, Jing Xu, and Y-Lan Boureau. 2022.
\newblock \href {https://doi.org/10.18653/v1/2022.acl-long.447}
  {{S}a{F}e{RD}ialogues: Taking feedback gracefully after conversational safety
  failures}.
\newblock In \emph{Proceedings of the 60th Annual Meeting of the Association
  for Computational Linguistics (Volume 1: Long Papers)}, pages 6462--6481,
  Dublin, Ireland. Association for Computational Linguistics.

\bibitem[{van~den Oord et~al.(2019)van~den Oord, Li, and
  Vinyals}]{oord2019representation}
Aaron van~den Oord, Yazhe Li, and Oriol Vinyals. 2019.
\newblock \href {http://arxiv.org/abs/1807.03748} {Representation learning with
  contrastive predictive coding}.

\bibitem[{Wolf et~al.(2020)Wolf, Debut, Sanh, Chaumond, Delangue, Moi, Cistac,
  Rault, Louf, Funtowicz, Davison, Shleifer, von Platen, Ma, Jernite, Plu, Xu,
  Le~Scao, Gugger, Drame, Lhoest, and Rush}]{wolf-etal-2020-transformers}
Thomas Wolf, Lysandre Debut, Victor Sanh, Julien Chaumond, Clement Delangue,
  Anthony Moi, Pierric Cistac, Tim Rault, Remi Louf, Morgan Funtowicz, Joe
  Davison, Sam Shleifer, Patrick von Platen, Clara Ma, Yacine Jernite, Julien
  Plu, Canwen Xu, Teven Le~Scao, Sylvain Gugger, Mariama Drame, Quentin Lhoest,
  and Alexander Rush. 2020.
\newblock \href {https://doi.org/10.18653/v1/2020.emnlp-demos.6} {Transformers:
  State-of-the-art natural language processing}.
\newblock In \emph{Proceedings of the 2020 Conference on Empirical Methods in
  Natural Language Processing: System Demonstrations}, pages 38--45, Online.
  Association for Computational Linguistics.

\bibitem[{Xu et~al.(2021{\natexlab{a}})Xu, Ju, Li, Boureau, Weston, and
  Dinan}]{xu-etal-2021-bot}
Jing Xu, Da~Ju, Margaret Li, Y-Lan Boureau, Jason Weston, and Emily Dinan.
  2021{\natexlab{a}}.
\newblock \href {https://doi.org/10.18653/v1/2021.naacl-main.235}
  {Bot-adversarial dialogue for safe conversational agents}.
\newblock In \emph{Proceedings of the 2021 Conference of the North American
  Chapter of the Association for Computational Linguistics: Human Language
  Technologies}, pages 2950--2968, Online. Association for Computational
  Linguistics.

\bibitem[{Xu et~al.(2021{\natexlab{b}})Xu, Ju, Li, Boureau, Weston, and
  Dinan}]{xu2021recipes}
Jing Xu, Da~Ju, Margaret Li, Y-Lan Boureau, Jason Weston, and Emily Dinan.
  2021{\natexlab{b}}.
\newblock \href {http://arxiv.org/abs/2010.07079} {Recipes for safety in
  open-domain chatbots}.

\bibitem[{Zhang et~al.(2018)Zhang, Dinan, Urbanek, Szlam, Kiela, and
  Weston}]{zhang2018personalizing}
Saizheng Zhang, Emily Dinan, Jack Urbanek, Arthur Szlam, Douwe Kiela, and Jason
  Weston. 2018.
\newblock \href {http://arxiv.org/abs/1801.07243} {Personalizing dialogue
  agents: I have a dog, do you have pets too?}

\bibitem[{Ziems et~al.(2022)Ziems, Yu, Wang, Halevy, and
  Yang}]{ziems-etal-2022-moral}
Caleb Ziems, Jane Yu, Yi-Chia Wang, Alon Halevy, and Diyi Yang. 2022.
\newblock \href {https://doi.org/10.18653/v1/2022.acl-long.261} {The moral
  integrity corpus: A benchmark for ethical dialogue systems}.
\newblock In \emph{Proceedings of the 60th Annual Meeting of the Association
  for Computational Linguistics (Volume 1: Long Papers)}, pages 3755--3773,
  Dublin, Ireland. Association for Computational Linguistics.

\end{thebibliography}
\bibliographystyle{acl_natbib}

\appendix
\section{Datasets} \label{ap:datasets}
In this study, we will utilize two different classes of datasets. The first class $\clubsuit$ comprises datasets encompassing harmful conversation scenarios and corresponding mitigation strategies. The second class $\heartsuit$ consists of general-purpose chitchat datasets, which allows us to explore how language models can generate harmful or socially inept conversations. Below are the details:
\begin{itemize}[noitemsep,nolistsep, leftmargin=*]
\item \textbf{\textsc{Moral Integrity Copus(MIC)}$\clubsuit$}: \cite{ziems-etal-2022-moral} captures the moral
assumptions of 38k prompt-reply pairs, using
99k distinct Rules of Thumb (RoTs). Each
RoT reflects a particular moral conviction
that can explain why a chatbot’s reply may
appear acceptable or problematic.
\item \textbf{\textsc{ProsocialDialog}$\clubsuit$}:\cite{kim-etal-2022-prosocialdialog} contains responses that encourage prosocial behavior, grounded in commonsense social rules (i.e., rules of thumb or RoTs). Created via a human-AI collaborative framework, \textsc{ProsocialDialog} consists of 58K dialogues, with 331K utterances,
160K RoTs and 497K dialogue safety labels
accompanied by free-form rationales.
\item \textbf{DailyDialog}$\heartsuit$: \cite{li-etal-2017-dailydialog} The dialogues in the dataset reflect our daily communication way and cover various topics about our daily life. This dataset contains 13,118 multi-turn dialogues.
\item \textbf{Empathetic Dialogs}$\heartsuit$: \cite{rashkin-etal-2019-towards} is a novel
dataset of 25k conversations grounded in emotional situations.
\item \textbf{PersonaChat}$\heartsuit$: \cite{zhang2018personalizing} The dataset consists of 8939 complete dialogues for training, 1000 for validation, and 968 for testing. 
\item \textbf{Blended Skill Talk(BST)}$\heartsuit$: \cite{smith2020together} Engaging, knowledgeable, and empathetic are desirable general qualities in a conversational agent. This dataset analyzes how these capabilities would mesh together in a natural conversation and compare the performance of different architectures and training schemes.
\end{itemize}

\section{Natural occurrence of socially inappropriate situations}

\begin{figure}[h]
\centering
\includegraphics[width=0.4\textwidth]{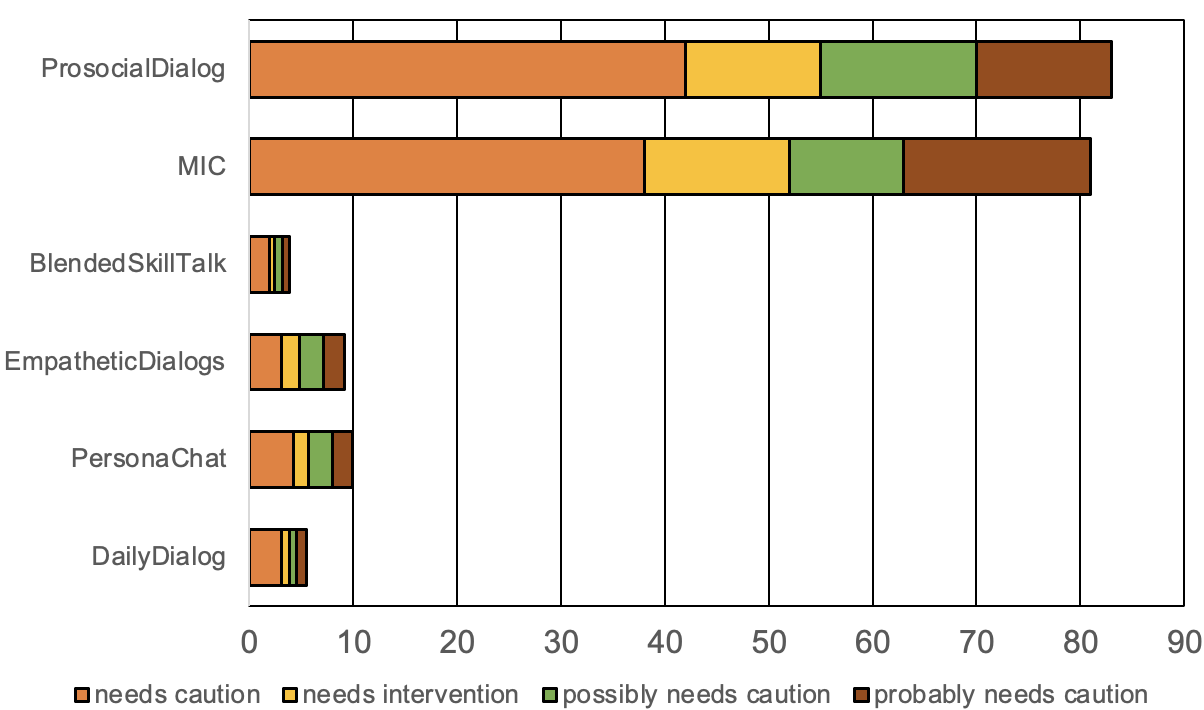}
\caption{Different percentages of unsocial content across multiple datasets. The definitions of each category are taken from the ProsocialDialog dataset and explained in \S \ref{sec:ap_labels}}
\label{fig:data_audit}
\end{figure}

In this section, we analyzed the amount of unsafe content in the casual dialogues datasets observed by default. Given the context (last turn), we classified each of the utterances in the dataset, given the context(prior turn), using a classifier described in \S \ref{sec:ap_clf}. As seen in Figure \ref{fig:data_audit}, an average $\sim$ 4-10\% of the data is classified as not casual. The hypothesis is that utterances that need extra caution or intervention can force the generative models to produce unsafe responses, disrupting the flow of the conversation and breaking the user’s trust.

\section{Dialog Safety Labels} \label{sec:ap_labels}
\begin{itemize}[noitemsep,nolistsep, leftmargin=*]
\item \textbf{Needs Intervention}: This pertains to instances where the utterances go beyond being problematic and necessitate human intervention for prosocial actions. Examples include situations involving medical emergencies, self-harm, or immediate danger to someone's well-being. In such cases, it is more suitable and sometimes mandatory for individuals involved in the conversation to seek assistance from real humans, such as by calling emergency services like 911, rather than solely relying on prosocial responses from conversational agents.
\item \textbf{Needs Caution}: describes utterances and situations that are potentially problematic, unethical, rude, toxic, or biased and may require caution to respond prosocially. The fine-grained labels for dialogues that needed caution are borrowed from the setting used in the \textsc{ProsocialDialog} dataset. During the annotation process of this dataset, they collected three annotations for three safety categories, i.e. (1) \textsc{Casual} (2) \textsc{Needs Caution} (3) \textsc{Needs Intervention}. Now, \textsc{Possibly Needs Caution}, \textsc{Probably Needs Caution} and \textsc{Needs Caution} refer to one, two, and three votes for ‘Needs Caution’ without any votes for ‘Needs Intervention’, respectively. So, the order of cases that needs more caution is like this:
\textsc{Needs Caution} > \textsc{Probably Needs Caution} > \textsc{Possibly Needs Caution}.
\end{itemize}

\section{Dialog Safety Classifier} \label{sec:ap_clf}

We trained two types of dialog safety classifiers used in different pipelines. The first one is a generative classifier. Following \cite{prost-2022-integrating}, we trained an encoder-decoder model(T5-base) to generate the safety label and RoT jointly. The base model was initialized with fine-tuned on Delhpi \cite{jiang2022machines} commonsense norm databank. Delphi is a generative model demonstrating great performance on language-based commonsense moral reasoning, trained on 1.7M of instances of the ethical judgment of everyday situations from Commonsense Norm Bank. We evaluate this first version of our safety classifier on \textsc{ProsocialDialog} validation and test sets. The results were mostly similar to the original paper. 76.6 \% validation accuracy was observed and 76.7 \% on test set. 

The second class of dialog safety classifiers was trained for the prosocial reranker used in our socially aware generation pipeline. In this classifier, we do binary classification, i.e., it is social or not social. This classifier has two types of architecture, it can do sentence pair classification(used in training), which is trained using a cross-encoder \cite{reimers-gurevych-2019-sentence} style network. Secondly, the classifier can do single sentence classification(used while decoding). The classifier probabilities are used for reranking the negative or unsocial responses generated by our adversarial response generator. We follow the same fine-tuning sequence as in the previous classifier. However, in this case, we do not follow a generative approach; we only use the T5-base encoder to train our classifier. The classification accuracy on \textsc{ProsocialDialog} test was 79.2 \%. Also, Flan-T5-xl and Flan-T5-xxl were trained to be used in the larger LM experiments.

\section{Rule of Thumb(RoT) Generator} \label{sec:ap_rot}

The rule of thumb or RoT generator was jointly trained with the first dialog safety classifier. The details of hyperparameters are as follows:

\begin{itemize}[noitemsep,nolistsep, leftmargin=*]
\item Base model: same as the main model(T5-base, COSMO, etc)

\item Dataset: ProsocialDialog.

\item Batch size: 8-2 (Varies depending on the model size)

\item Max context length: 128

\item Max training epochs: 10

\item Learning rate: 1.00E-05

\item Optimizer: Adam

\item Greedy decoding is used during inference.

\end{itemize}

\begin{table}[htb]
\centering
\scalebox{0.6}{%
\begin{tabular}{@{}l|rr@{}}
\toprule
\textbf{Model}                  & \multicolumn{1}{l}{\textbf{BLEU-4}} & \multicolumn{1}{l}{\textbf{PPL}} \\ \midrule
Canary(Delphi)                  & 16.5                                & 5.3                              \\
Ours(Only context)              & \textbf{19.7}                       & \textbf{4.1}                     \\
Ours(Only context and response) & 20.08*                              & 4.1                              \\ \bottomrule
\end{tabular}}
\caption{Performance of our RoT generator as compared to $\mathrm{Canary}$}
\label{tab:rot_model_performance}
\end{table}

The performance of a model trained on based T5-large is shown in Table \ref{tab:rot_model_performance}.  Adding control tokens while generating RoTs prove to be an effective strategy. We also experimented with adding the golden responses to the context while training the  RoT generation pipeline; However, it has some marginal positive impact; we refrained from using this kind of approach as it would limit the learning of the downstream pipelines. 






\section{Baselines} \label{sec:ap_baseline}

\begin{itemize}[noitemsep,nolistsep, leftmargin=*]
\item \textbf{Prost} \cite{kim-etal-2022-prosocialdialog}: Prosocial Transformer or Prost is trained on \textsc{ProsocialDialog} dataset using BlenderBot 2.7B as its backbone.  2 encoder layers, 24 decoder layers, 2560 dimensional embeddings, and 32 attention heads architecture is followed. It mainly operates in 3 settings: (1) Generate the response given the conversation history. (2) Generate the response and RoT given the conversation history. (3) Generate the response given the conversation history and golden RoT.
\item \textbf{\textsc{Dexperts}}\cite{liu-etal-2021-dexperts}:  \textsc{Dexperts}: Decoding-time Experts, a decoding time method for controlled text generation that combines a pre-trained language model with “expert” LMs and/or “anti-expert” LMs in a product of experts. Intuitively, under
the ensemble, tokens only get high probability if they are considered likely by the experts and unlikely by the anti-experts.  The product-of-experts ensemble is given by:

{
\small
\begin{equation}
    P(X_t|x_{<t}) = \mathrm{softmax}(\mathbf{z}_t+\alpha(\mathbf{z}^+_t-\mathbf{z}^-_t))
\end{equation}
}

Where $P(X_t|x_{<t})$ is the probability of generating $X_t$ given $x_{<t}$, $\mathbf{z}_t$ is the logit of $t$-th token from the base model, $\mathbf{z}^+_t$ is the logit of $t$-th token from the expert model and $\mathbf{z}^-_t$ is the logit of $t$-th token from the anti-expert model. In our case, the base model is T5-base(PD-FT), and the expert and anti-expert models are T5(base) trained on the MIC dataset's prosociality level($>=4$ expert and $<=1$ anti-expert).

\item \textbf{Contrastive Decoding(CD)}\cite{li-etal-2023-contrastive}: this idea is an extension of \textsc{Dexperts}, here a contrastive objective is defined that returns the difference between the likelihood under an expert and amateur model. The ensemble is defined as:

{
\small
\begin{equation}
    P(X_t|x_{<t}) = \mathrm{softmax}(\mathbf{z}^{\mathrm{exp}}_t-\mathbf{z}^{\mathrm{ama}}_t)
\end{equation}
}

Where $P(X_t|x_{<t})$ is the probability of generating $X_t$ given $x_{<t}$, $\mathbf{z}^{exp}_t$ is the logit of $t$-th token from the expert model and $\mathbf{z}^{ama}_t$ is the logit of $t$-th token from the amateur model. The expert model is the same as \textbf{\textsc{T5}-base(PD-FT)}, and the amateur model is the same as the anti-expert model explained in  \textbf{\textsc{Dexperts}}.

\end{itemize}

\section{\textsc{Cosmo}} \label{sec:ap_cosmo}

\textsc{Cosmo} \cite{kim2023soda} is a generalizable conversation model that is significantly more natural and consistent on unseen datasets than best-performing conversation models (e.g., GODEL, BlenderBot-1, Koala, Vicuna). \textsc{Cosmo} is trained on SODA, a million-scale high-quality social dialogue dataset, and \textsc{ProsocialDialogs} dataset. It has two versions \textsc{Cosmo}(3B) and \textsc{Cosmo}(11B); the base models used here are derived from T5X library. More details can be found in the paper.

\section{Implementation Details}

All the models in our pipeline, including the base and final, are implemented using the Pytorch Huggingface Transformers library\cite{wolf-etal-2020-transformers} and Deepspeed \cite{deepspeed}\footnote{\href{https://huggingface.co/docs/transformers/main_classes/deepspeed}{https://huggingface.co/docs/transformers\\/main\_classes/deepspeed}}. The following configuration was best performing for the base, and the final models are shown in Table \ref{tab:hyp_base} and \ref{tab:hyp_final}. The smaller models were trained in two NVIDIA A5000 GPUs; the average running time for the base models was 2 hours, and for the final models was $\sim$ 5-7 hours. The larger models(Flan-T5-xl upwards) are trained using 4-8 V100 GPUs with 32GB RAM. The average runtime for base models is 1.5 hours for the base model and $\sim$ 4 hours for the final model.  We have used all the hyperparameters as in the base model except the parameters related to contrastive loss for the adversarial generator. 

\begin{table}[htb]
\centering
\scalebox{0.8}{%
\begin{tabular}{@{}l|r@{}}
\toprule
\textbf{Hyper-parameter} & \multicolumn{1}{l}{\textbf{Value}} \\ \midrule
base pre-trained model  & \multicolumn{1}{l}{t5-base}        \\
batch size              & 8                                  \\
max context length              & 128                                 \\
\# training epochs      & 10                                 \\
learning rate           & 3.00E-05                           \\
alpha                   & 0.5                                \\
oracle function     & BLEU                                \\
max length              & 60                                 \\
min length              & 5                                  \\
diversity penalty       & 2                                  \\
max negative sample \#  & 12                                 \\
no-repeat ngram         & 4                                  \\
early stop             & \multicolumn{1}{c}{TRUE}           \\ \bottomrule
\end{tabular}}
\caption{Base model hyper-parameters(small LM)}
\label{tab:hyp_base}
\end{table}

\begin{table}[htb]
\centering
\scalebox{0.8}{%
\begin{tabular}{@{}l|r@{}}
\toprule
\textbf{Hyper-parameter} & \multicolumn{1}{l}{\textbf{Value}} \\ \midrule
base pre-trained model  & \multicolumn{1}{l}{t5-base}        \\
batch size              & 4                                  \\
max context length              & 128                                 \\
\# training epochs      & 10                                 \\
learning rate           & 2.00E-05                           \\
alpha                   & 0.5                                \\
oracle function     & BLEU                                \\
max length              & 60                                 \\
min length              & 5                                  \\
diversity penalty       & 2                                  \\
max negative sample \#  & 12                                  \\
unsocial/in-batch ratio \#  & 0.75                                  \\
no-repeat ngram         & 4                                  \\
early stop             & \multicolumn{1}{c}{TRUE}           \\ \bottomrule
\end{tabular}}
\caption{Final model hyper-parameters(small LM)}
\label{tab:hyp_final}
\end{table}

\begin{table}[htb]
\centering
\scalebox{0.8}{%
\begin{tabular}{@{}l|r@{}}
\toprule
\textbf{Hyper-parameter} & \multicolumn{1}{l}{\textbf{Value}} \\ \midrule
r  & 16        \\
lora\_alpha              & 32                                \\
target\_modules      & "q", "v"                                 \\
lora\_dropout           & 0.05                           \\
bias                   & None                                  \\ \bottomrule
\end{tabular}}
\caption{LoRA hyperparameters}
\label{tab:hyp_lora}
\end{table}

\begin{table}[htb]
\centering
\scalebox{0.8}{%
\begin{tabular}{@{}l|r@{}}
\toprule
\textbf{Hyper-parameter} & \multicolumn{1}{l}{\textbf{Value}} \\ \midrule
base pre-trained model  & \multicolumn{1}{l}{A, B}        \\
batch size              & 2                                  \\
max context length              & 128                                 \\
\# training epochs      & 10                                 \\
learning rate           & 2.00E-05                           \\
alpha                   & 0.5                                \\
oracle function     & BLEU                                \\
max length              & 60                                 \\
min length              & 5                                  \\
diversity penalty       & 2                                  \\
max negative sample \#  & 8                                  \\
no-repeat ngram         & 4                                  \\
early stop             & \multicolumn{1}{c}{TRUE}           \\ \bottomrule
\end{tabular}}
\caption{Base model hyper-parameters(large LM), A=\texttt{Flan-T5}(xl or xxl), B=\textsc{Cosmo}(3B or 11B), n\_gpus depend on the size of the model, 4 for 3B and 8 for 11B}
\label{tab:hyp_base}
\end{table}

\begin{table}[htb]
\centering
\scalebox{0.8}{%
\begin{tabular}{@{}l|r@{}}
\toprule
\textbf{Hyper-parameter} & \multicolumn{1}{l}{\textbf{Value}} \\ \midrule
base pre-trained model  & \multicolumn{1}{l}{A, B}        \\
batch size              & 1                                  \\
max context length              & 128                                 \\
\# training epochs      & 10                                 \\
learning rate           & 2.00E-05                           \\
alpha                   & 0.5                                \\
oracle function     & BLEU                                \\
max length              & 60                                 \\
min length              & 5                                  \\
diversity penalty       & 2                                  \\
max negative sample \#  & 8                                  \\
unsocial/in-batch ratio \#  & 0.75                                  \\
no-repeat ngram         & 4                                  \\
early stop             & \multicolumn{1}{c}{TRUE}           \\ \bottomrule
\end{tabular}}
\caption{Final model hyper-parameters(large LM), A=\texttt{Flan-T5}(xl or xxl), B=\textsc{Cosmo}(3B or 11B), n\_gpus depend on the size of the model, 4 for 3B and 8 for 11B}
\label{tab:hyp_final}
\end{table}

\section{Human Evaluation Details} \label{sec:ap_human_eval}
Here, we detail the instructions given to the human evaluators. The evaluators were given the context, i.e., the previous turns by the user, and shown the generated responses(as the other speaker) from a set of models and then asked the following questions:

\begin{enumerate}
  \item \textit{Prosociality}: Which response better implies that the other speaker should behave prosocially, ethically, and follow social norms?
  \item \textit{Engaged}: Which response is more engaged, curious, or empathetic towards the other speaker?
  \item \textit{Respect}: Which response is more respectful,
kind, and polite towards the other speaker?
  \item \textit{Coherency}: Which response is more contextually relevant and coherent in the context of the conversation?
  \item \textit{Overall}: Which response is the
best/most suited given the full conversation?
\end{enumerate}

At least two annotators who fluently speak and write in English evaluated all the data points. Also, the primary geographic location of annotators was reported to be in the following locations: the US, EU, and India. The annotators were paid 10-15\$ an hour. Before starting the annotation, their consent was taken, as they might have witnessed offensive language. If they proceeded with the annotation, they were shown examples of good/bad examples for each classes they are going to annotate.

\begin{figure}[h]
\centering
\includegraphics[width=0.4\textwidth]{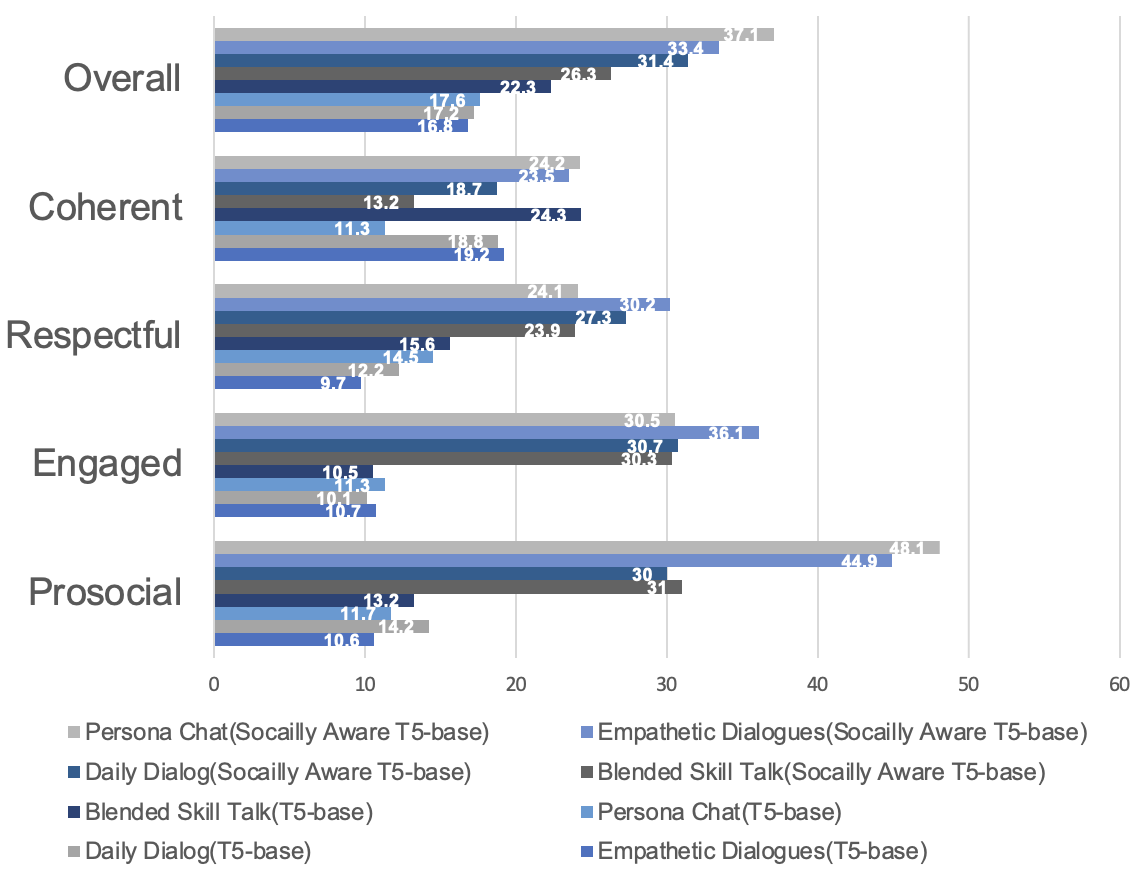}
\caption{Larger-scale(400 samples) human evaluation results on chit-chat dialog datasets.}
\label{fig:human_eval_data}
\end{figure}

\subsection{Prompting Details} \label{sec:ap_prompt}

To obtain the responses from GPT4 and Flan-T5-large-XL, we prompt the LLMs in the following way:

$\texttt{Given this utterance by a user: }$
$\texttt{<Context> \textbackslash n} $  

$\texttt{And a social norm that needs }$
$\texttt{to be followed: <Social Norm>\textbackslash n}$

$\texttt{Generate a reply following }$
$\texttt{the social norm in one sentence.}$

\end{document}